\documentclass{article}

\PassOptionsToPackage{square, numbers, sort&compress}{natbib}
\usepackage{iclr2025_conference,times}

\usepackage[utf8]{inputenc} %
\usepackage[T1]{fontenc}    %
\usepackage{hyperref}       %
\usepackage{url}            %
\usepackage{booktabs}       %
\usepackage{amsfonts}       %
\usepackage{nicefrac}       %
\usepackage{microtype}      %
\usepackage{xcolor}         %
\usepackage{graphicx}       %
\usepackage{amsmath}        %
\usepackage[normalem]{ulem} %

\DeclareMathOperator*{\argmin}{arg\,min} %

\definecolor{green}{rgb}{0.0, 0.64, 0.0}
\definecolor{red}{rgb}{1.0, 0.13, 0.32}
\definecolor{blue}{rgb}{0.06, 0.2, 0.65}
\definecolor{darkgreen}{rgb}{0,0.5,0}
\definecolor{darkblue}{rgb}{0,0,0.6}
\definecolor{purple}{rgb}{0.4,.2,0.7}
\definecolor{magenta}{rgb}{1.0,0.0,1.0}
\definecolor{orange}{rgb}{1.0,0.65,0}
\definecolor{darkorange}{rgb}{1.0,0.40,0}

\title{The Unreasonable Ineffectiveness of the Deeper Layers}

\author{%
Andrey Gromov\thanks{Co-first authors; please direct correspondence to the union of \{\texttt{gromovand@meta.com}, \texttt{kushaltirumala99@gmail.com}, \texttt{drob@mit.edu}\}. \\ 
} \\
Meta FAIR \& UMD \\ 
\And
Kushal Tirumala$^*$ \\
Meta FAIR \\
\And
Hassan Shapourian \\
Cisco
\And
Paolo Glorioso \\
Zyphra \\
\AND
Daniel A. Roberts%
\\
MIT \& Sequoia Capital
}

\iclrfinalcopy
\begin{document}
\maketitle

\begin{abstract}
How is knowledge stored in an LLM's weights?
We study this via layer pruning: if removing a certain layer does not affect model performance in 
common question-answering benchmarks,
then the weights in that layer are not necessary for storing the knowledge needed to answer those questions.
To find these 
unnecessary parameters,
we identify the optimal block of layers to prune 
by considering 
similarity 
across
layers;
then,
to ``heal'' the damage, we 
perform a small amount of finetuning.
Surprisingly, with this method we find minimal degradation of performance 
until after a large fraction (up to half) of the layers
are removed for some common open-weight models.
From a scientific perspective, 
the robustness of these LLMs to the 
deletion of 
layers
implies either that current pretraining methods 
are
not properly 
leveraging
the
parameters in the deeper layers of the network
or that the shallow layers play a critical role in storing knowledge.
For our study, we use
parameter-efficient finetuning (PEFT) methods, 
specifically quantization and Low Rank Adapters (QLoRA), 
such that each of our experiments can be performed on a single 40GB A100 GPU.

\end{abstract}

\section{Introduction}
\label{sec:intro}

\begin{figure}[t]
\begin{center}
\centerline{\includegraphics[width=1.0\columnwidth]
{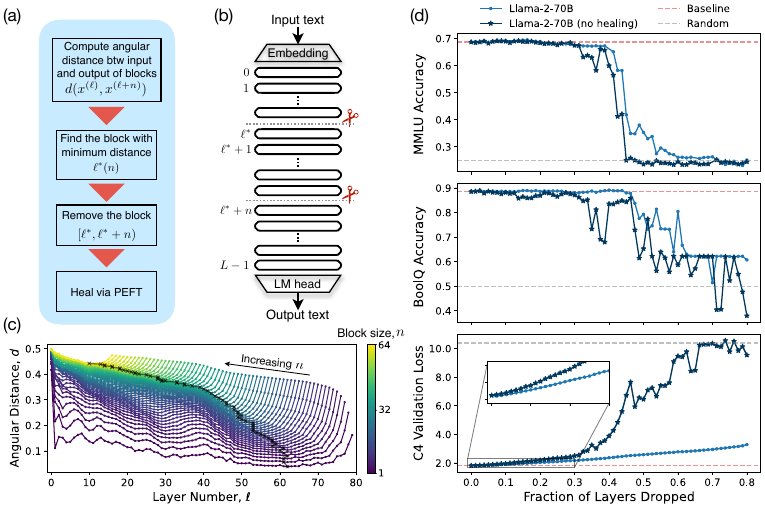}}
\caption{Overview of our layer-pruning strategy and example results: 
\emph{(a)} a flowchart describing the algorithm: if removing $n$ layers, we find the layer, $\ell^*$, that minimizes the angular distance, $d$, between layers $\ell$ and $\ell\! +\! n$; we then remove the $n$ layers beginning with layer $\ell^*$; finally, if necessary, we can ``heal'' the damage with a small amount of (parameter-efficient) finetuning.
\emph{(b)} a schematic depicting the removal of $n$ total layers, indexed from $\ell^*\!$ to $\ell^* \!\!+ \!n \!-\! 1$. %
\emph{(c)} angular distance, $d$, between different numbers of layers, $n$, vs. the layer number, $\ell$, that indexes the beginning of the block of $n$; the bottom curve (darkest purple) represents $n=1$, while the top curve (lightest yellow) represents $n=64$; the black line traces $\ell^*(n)$, the minimum of the angular distance across the different sized layer blocks. 
\emph{(d)} results of pruning Llama-2-70B with healing (light blue) and without healing (dark blue) as a function of the fraction of layers removed: the top (middle) panel gives the accuracy on the MMLU (BoolQ) question-answering benchmark, while the bottom panel the autoregressive loss on a subset of the C4 validation set;
here, the dashed red lines (dashed gray lines) indicate the accuracy or loss of the original unpruned model (of random guessing);
these plots illustrate that typical behavior we find in which there are sharp transitions in performance for the accuracy of question-answering tasks (here between 40\%-50\% pruning fraction), but continuity and very slow growth in the healed loss (light blue) up to at least to 80\% pruning fraction.
}
\label{fig:schematic-llama2-70b}
\end{center}
\end{figure}

In this work
we study
a very simple pruning strategy using
open-weight LLMs.
In particular, we develop a 
method
that 
uses the similarity between the representations at different layers to
identify the optimal layers to prune for a given pruning fraction; 
then,
after removing these layers we ``heal'' the 
pruning-induced 
mismatch 
with a small amount of fine tuning (using QLoRA).
Our main result is that we can remove a substantial fraction of the \emph{deepest layers} from models with minimal degradation in downstream question-answering  benchmarks. 
For example, 
for Llama-2-70B \citep{touvron2023llama2} 
we can eliminate up to roughly \emph{half} of the layers
before the performance collapses.
An overview of our strategy and the results of pruning Llama-2-70B are shown in Figure~\ref{fig:schematic-llama2-70b}.

Our intuition for dropping layers comes from considering the residual structure of the transformer architecture. In more detail, the output of the final layer can be decomposed as a sum over the outputs of all the model layers plus the 
embedded input.
If such a sum 
had numerous
and 
independent terms, then removing a handful of 
them
should not significantly change the output. However, since the terms are not independent 
-- each layer is 
input to the following layer -- 
we should expect to be able to remove terms if the residual contribution from a particular layer is small. In other words, if the output of each layer does not change too much from layer to layer.\footnote{This is strongly suggested 
by 
``lens'' investigations 
that studied the evolution of the token distribution 
as a function of layer index
such as
the ``logit lens'' \citep{logit_lens_2020} and the
``tuned lens'' \citep{belrose2023eliciting}.
A separate line of
reasoning along these lines previously inspired neural ODEs \citep{chen2018neural},
and led \citet{yang2023tensor}
to argue
that 
ideally representation should change substantially 
from layer to layer 
in order to most
effectively make use of the parameters 
of a network.
}

In conjunction with our layer pruning, we investigate
the similarity of layer representations at different separations
and find broadly that deeper layers are qualitatively more similar to neighboring layers than shallow layers (with the exception of the very final layer).
This suggests an even simpler pruning strategy: remove layers beginning at the penultimate layer and proceed from deep to shallow until the desired number of layers have been removed.
In this case, we find that,
after healing the damage with a small amount of QLoRA finetuning,
we can achieve performance that nearly matches the more involved similarity-informed layer pruning strategy.
The effectiveness of this method is 
evidence that LLMs might not properly leverage the parameters in the deeper layers of the network.

That said, while question-answering (QA) benchmarks such as MMLU and BoolQ are robust to a large amount of layer pruning, other measures of performance are not: if we look at the loss on next-token predictions for an IID dataset (C4 validation set), we find that the model is smoothly damaged in proportion to the fraction of the number of layers pruned. Since perplexity typically correlates strongly with downstream metrics, this naturally begs the question: which tasks are less robust than QA benchmarks to pruning? As part of our final discussion, we explore reasoning related tasks (GSM8k and HellaSwag) and see that they are harmed by any amount of pruning. Altogether, this leads to the following accounting of state: the shallow layers likely play a critical role in the storing of knowledge and retrieving of information, while the deeper layers are important for higher-level computations such as mathematical reasoning.

The structure of this paper is as follows. In \S\ref{sec:review}, we first perform a literature review of both practical post-training strategies and science-of-deep-learning investigations that motivate our work. 
Then, in \S\ref{sec:method}, we give intuition for our layer pruning strategy
and explain our method in detail,
while in \S\ref{sec:results} we
iterate over all our experimental results. Finally, we conclude in \S\ref{sec:conclusion} 
by exploring
tasks beyond QA benchmarks, such as reasoning, and
highlighting directions of future work. Specific model, finetuning, dataset, and evaluation details can be found in Appendix~\ref{app:healing-procedure}, 
and
evaluation ablations can be found in 
Appendix~\ref{app:hyperpara}.

\section{Literature Review}
\label{sec:review}

Pruning for neural networks has a long history \citep{NIPS1989_6c9882bb, NIPS1992_303ed4c6}: while initial work focused on \emph{unstructured pruning} \citep{han2015learning, chen2015compressing, srinivas2015data}, \emph{structured pruning} techniques were developed to make sparse networks more efficient \citep{li2016pruning,wen2016learning, hu2016network,he2017channel,huang2018condensenet, murray2015auto, see2016compression, kim2016sequence}. Recent work, of course, focused on structured pruning of transformers \citep{voita2019analyzing, michel2019sixteen, kim2020fastformers,fan2019reducing, zhang2020accelerating, fan2021layer,jha2023large,sajjad2023effect,liu2023comflp,hou2020dynabert,sharma2023truth,ashkboos2024sliceGPT,xia2022structured, lagunas2021block,men2024shortgpt}. Our work focuses on pruning the layers of decoder-only GPT style open-weight \emph{large} language models after they've been pretrained. For an extended literature review, please see Appendix~\ref{app:extended-lit-review}.

\section{Method}
\label{sec:method}

In this section, we give intuition for why we think 
layer pruning works
(\S\ref{subsec:intuition})
and then we explain our method in detail (\S\ref{subsec:layer-pruning-algo}).

\subsection{Intuition}\label{subsec:intuition}

Our intuition for layer dropping comes from thinking about 
the 
representations
as a slowly changing function of layer index. In particular, the
layer-to-layer evolution of representations for a transformer is given by a \emph{residual} iteration equation
\begin{equation}\label{eq:iteration-normal}
x^{(\ell + 1)} = x^{(\ell)} +  f(x^{(\ell)}, \theta^{(\ell)}) \, ,
\end{equation}
where $(x^{(\ell)}$, $\theta^{(\ell)})$, respectively, are the multi-dimensional input and parameter vectors for layer $\ell$, and $f(x, \theta)$ describes the transformation of 
one
multi-head self-attention \emph{and} MLP
layer block. As for any residual network, if we unroll this iteration, we see that after $L$ total layers the output is described as a sum over the transformations of all the layers
\begin{equation}\label{eq:unrolled}
x^{(L)} = x^{(0)} +  \sum_{\ell=0}^{L-1} f(x^{(\ell)}, \theta^{(\ell)}) \, .
\end{equation}
If the terms in the sum were 
\emph{numerous}, ($L \gg 1$), 
and
\emph{independent}, e.g. if the block functions were instead a function of the overall input as $f(x^{(0)}, \theta^{(\ell)})$, 
then
perhaps any particular contribution to the sum \eqref{eq:unrolled} could be neglected.

Of course, they are not at all independent:
if we delete layer $\ell-1$, then we must now 
connect
the old input to that layer, 
$x^{(\ell-1)}$,
into the block function of layer $\ell$ as
\begin{equation}\label{eq:iteration-deleted}
x^{(\ell + 1)} = x^{(\ell-1)} +  f(x^{(\ell-1)}, \theta^{(\ell)}) \, ,
\end{equation}
where, for clarity, we are not relabeling layers or inputs despite the deletion.
In general, such a \emph{mismatch} between the original input and new input should be very damaging for the network.
However, if, after some number of initial layers, the representations converge to a slowly changing function with respect to layer index,
\begin{equation}
x^{(\ell)} \approx x^{(\ell-1)} + \epsilon \, ,
\end{equation}
with $\epsilon  \ll x^{(\ell)}$ in some appropriate sense, then the effect of deleting a particular layer $\ell$, e.g. making the replacement $x^{(\ell)} \to x^{(\ell-1)}$ in going from \eqref{eq:iteration-normal} to \eqref{eq:iteration-deleted}, 
should
only change the representation in the subsequent layer, $x^{(\ell + 1)}$, by a small amount. 
Similarly, 
to
successfully prune the $n$  layers before layer $\ell$, i.e. those indexed
from $\ell- n, \ldots, \ell -1$, 
we'd %
want
that the input 
to the pruned block
should be very similar to the output of the pruned block:
\begin{equation}\label{eq:mismatch-n}
x^{(\ell)} \approx x^{(\ell-n)} + \epsilon \, .
\end{equation}

Regardless, any layer removal
has a cascading effect:
since post pruning $x^{(\ell + 1)}$ is computed by a different function than before, cf. \eqref{eq:iteration-normal} vs. \eqref{eq:iteration-deleted}, and since then $x^{(\ell + 1)}$ is directly or indirectly input to subsequent layers,
$\ell + 2, \ldots, L$,
deleting a shallow layer 
should have a much greater impact
than deleting a deeper layer.

From this, we have the following 
hypotheses
that we will test experimentally:
\begin{enumerate}
\item[\emph{(0)}] We should be able to prune layers of a residual network.
\item[\emph{(1)}] We should have greater success pruning deeper layers.
\item[\emph{(2)}] Blocks of layers 
we successfully prune should have outputs that are 
similar to their inputs.
\end{enumerate}
In the next subsection, \S\ref{subsec:layer-pruning-algo} we will explain the details of our pruning algorithm and in the following section, 
\S\ref{sec:results},
we will present experimental evidence for points~\emph{(0)-(2)}.

\subsection{Layer-pruning algorithm(s)}
\label{subsec:layer-pruning-algo}

Our principal layer pruning algorithm is very simple:
\begin{enumerate}
\item[0.] Pick a a number of layers to prune $n$.
\item Compute the angular distance 
$d(x^{(\ell)},x^{(\ell+n)})$, 
cf. \eqref{eq:arccos-sim} below,
between the input to layer $\ell$ and the input to layer $\ell +n$ on a neutral pretraining dataset or on a dataset representative of a downstream task of interest. 
\item Find the layer, $\ell^*$, that minimizes that distance:
\begin{equation}\label{eq:optimal-block}
\ell^\star(n) \equiv \argmin_\ell~ d(x^{(\ell)},x^{(\ell+n)}) \,.
\end{equation}
\item Drop layers $\ell^\star$ to $\ell^\star\!\!+\!n\!-\!1$; connect the old input to layer $\ell^\star$ to the old $(\ell^\star\!\!+\!n)$th layer block.\footnote{Layers are often contained in a data structure, such a \texttt{ModuleList} in \emph{PyTorch}, so to drop these layers we would simply define a new \texttt{ModuleList} that removes the 
layers from $\ell^\star$ to $\ell^\star + n - 1$.
}
\item (Optionally) heal the mismatch at layer $\ell^\star \!+ n$ with a small amount of fine tuning on a neutral pretraining dataset or particular dataset of interest.
\end{enumerate}
If fewer words inside of a figure are more helpful to you than the text in an enumerated list, then note that this algorithm is also depicted in panels (a)-(b) of Figure~\ref{fig:schematic-llama2-70b}.

Elaborating on the first step, the angular distance on a single sequence of length $T$ is given by
\begin{equation}\label{eq:arccos-sim}
d(x^{(\ell)},x^{(\ell+n)}) \equiv \frac 1\pi \arccos\left(\frac{x^{(\ell)}_T\cdot x^{(\ell+n)}_T}{\left|\!\left|x^{(\ell)}_T\right|\!\right| \left|\!\left|x^{(\ell+n)}_T\right|\!\right| } \right)\,,
\end{equation}
where the inner product is
over the hidden dimension of the model 
for the final token $T$ of the sequence,
$|\!|\cdot|\!|$ denotes the $L^2$-norm,
and the factor of $1/\pi$ is a 
convention.\footnote{Two comments: \emph{(i)}, we do not expect our choice of angular distance -- in lieu of any other reasonable metric, e.g., such as cosine similarity --
to be particular significant; and \emph{(ii)},
we chose to focus on the final token 
since, 
due to the causal attention mask,
its embedding is the only one that depends on the entire sequence.
} 
This distance should then be summed over a number of examples that is large enough to get a low-fluctuation estimate but overall should be quite small.

Elaborating on the ``optionality'' of the final step, we find that the near-lack of performance degradation on question-answering benchmarks, cf. Figure~\ref{fig:schematic-llama2-70b}(d) and others in \S\ref{subsec:transition}, can be extended to greater pruning fractions with a small amount of finetuning. Depending on resource constraints and intended application of the pruned model, this may not be necessary. However, 
the healing procedure 
does have
a substantial impact on perplexity, cf. Figure~\ref{fig:schematic-llama2-70b}(d) and others in \S\ref{subsec:other_measures}.

For both the angular distance measuring and the healing,
if the ultimate goal is to supervise finetune (SFT) a model for a downstream task, it could be useful to evaluate the distance of a sample from that dataset and
then
combine the healing process with the SFT. 
In contrast, for the greatest generality, it's most natural to measure distance and heal with a pretraining dataset that approximates the statistics under which the model was originally pretrained.

Finally, we also investigated an even simpler pruning strategy inspired 
by
analyzing 
the
angular distances across different model families:
drop
the deepest layers, 
excluding
the final layer before the LLM head, and then (\emph{non-optionally}) heal the damage. For complete clarity, 
this means that 
if 
we are 
pruning $n$ layers from an $L$-layer model,
then we would remove layers $(L - n)$ to $(L- 1)$, inclusive.

\section{Results}
\label{sec:results}

In this section, we demonstrate the effectiveness of our pruning strategy
on different 
question-answering (QA) benchmarks
and highlight
a robust 
pruning-driven transition in performance 
(\S\ref{subsec:transition}),
while,
in contrast, 
we 
find
that 
the autoregressive perplexities of the healed pruned models
are
continuous across their transition points (\S\ref{subsec:other_measures});
then, after comparing the similarity statistics between different layers across model sizes and families (\S\ref{subsec:activation}), we contrast our principal similarity-informed pruning strategy with a simpler remove-the-deepest-layers strategy (\S\ref{subsec:other_patterns}).

For our experiments, 
we pruned a wide variety of large-scale LLMs from 2.7B to 70B parameters spanning 32 to 80 total unpruned layers. Specifically, we used models in
the Llama-2 family  
\citep{touvron2023llama2},
the Qwen family \citep{bai2023qwen},
Mistral-7B \citep{jiang2023mistral},
and  Phi-2 \citep{phi2}.
For these models, we
executed
the ``healing'' step using
QLoRA \citep{dettmers2023qlora}: our models were quantized to 4-bit precision and then finetuned, using QLoRA for efficient training, 
on 
either
164M or 328M 
tokens 
from
the Colossal Clean Crawled Corpus (C4) \citep{raffel2020exploring}, a common pretraining dataset. 
As a result,
\emph{each experiment of ours can be performed on a single 40GB A$100$ GPU}. 
For our QA evals, we used
Massive Multitask Language Understanding (MMLU) \citep{hendrycks2020measuring}, 
a common world-knowledge and problem solving benchmark,
and
BoolQ \citep{clark2019boolq}, a common 
yes/no 
reading comprehension
benchmark 
where the answer has to be inferred from the text itself.
The specifics 
of our 
models,
healing
procedure, dataset choices, and evaluation details 
can be found across Appendix~\ref{app:healing-procedure};
ablations of different hyperparameter choices can be found across
Appendix~\ref{app:hyperpara}.

\subsection{Accuracy on QA benchmarks}
\label{subsec:transition}

\begin{figure}[t]
\begin{center}
\centerline{\includegraphics[width=1.0\columnwidth]{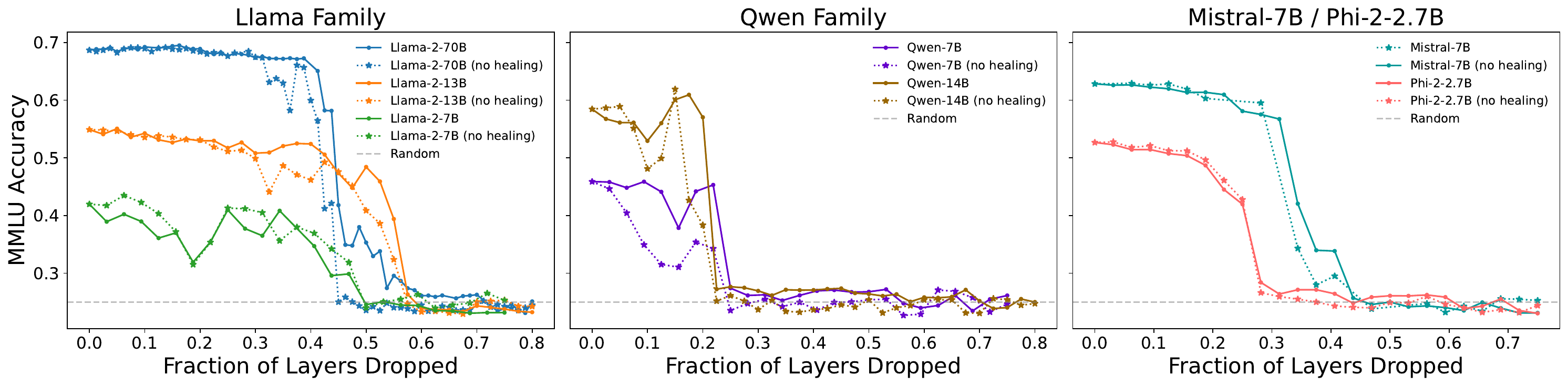}}
\caption{MMLU accuracy (5-shot) vs. fraction of layers 
dropped
for different model families. 
(\emph{Left:} Llama-2 family; \emph{Middle:} Qwen family; \emph{Right:} Mistral-7B and Phi-2.)
The 
solid lines
represent 
performance
after dropping layers and healing, 
dotted lines
show 
performance after dropping layers only (no healing),
and the dashed gray line is the score for guessing randomly.
For these models, healing leads to modest improvements, and performances 
are quite robust 
until 20\%-55\% pruning fractions, depending on model family and size, at which point they transitions to random guessing.
}
\label{fig:main-results-pruning}
\end{center}
\end{figure}

Our first set of results are shown in
Figure~\ref{fig:main-results-pruning}, 
where
we plot 
$5$-shot
MMLU accuracy as a function of the fraction of
layers removed: 
in the left panel
we present the Llama-2 family, 
in the middle panel
we present models from the Qwen family,
and in the right panel 
we show Mistral-7B and Phi-2.
In order to better compare models of different total %
number of layers,
in these plots we opted to
normalize the $x$-axis by the
fraction of layers removed (rather than the absolute number of layers removed).
Note that since MMLU contains multiple choice questions with four possible responses, the expected accuracy of random guessing is 25\%.

Importantly, we see a characteristic flat region of robust performance followed by a sharp transition to random accuracy at a pruning fraction around 45\%-55\%
for models in the Llama-2 family, 35\% for Mistral 7B,  25\% for Phi-2, and 20\% for models from the Qwen family. 
This implies that the essential knowledge required to achieve a model's top score 
isn't removed by significant layer
removal
-- even 
though the fraction can be quite
large(!) -- 
until eventually 
that knowledge is lost
at a critical model-dependent threshold.\footnote{This effect
is rather
robust
to choice of QA benchmark: in Figure~\ref{fig:appendix_boolq_acc} we plot the average 0-shot BoolQ accuracy for our model families and observe analogous behavior.}
Contrasting the curves with and without healing, we see that finetuning offers a modest improvement by better preserving the unpruned performance and pushing the phase transition to random guessing to slightly larger pruning fractions.

Broadly we see that layer pruning is more robust for the larger and deeper models, e.g. Llama-2-13B and Llama-2-70B, which we hypothesize could be related to the fact that either the smaller models are more overtrained, making parameters less redundant, or that the deeper models can afford to lose more layers in an absolute sense. Also, the Qwen family is strange, a fact we will further elaborate on in \S\ref{subsec:activation}.

\subsection{Loss 
on 
next-token predictions}
\label{subsec:other_measures}

In this section, 
we 
look at
the effect of layer pruning on 
the pretraining optimization objective
-- the cross-entropy loss of next-token prediction -- 
when evaluated on a subset of
the C4 validation dataset.\footnote{We make sure that none of the validation data are seen during the healing 
stage.}
In order to have a fair comparison across models with different sized vocabularies $V$, we normalize the loss by $\log V$, which corresponds to the loss of sampling tokens randomly with uniform probability.
(See Appendix~\ref{app:evaluation_details} for more details.)

\begin{figure}[t]
\begin{center}
\centerline{\includegraphics[width=1.0\columnwidth]{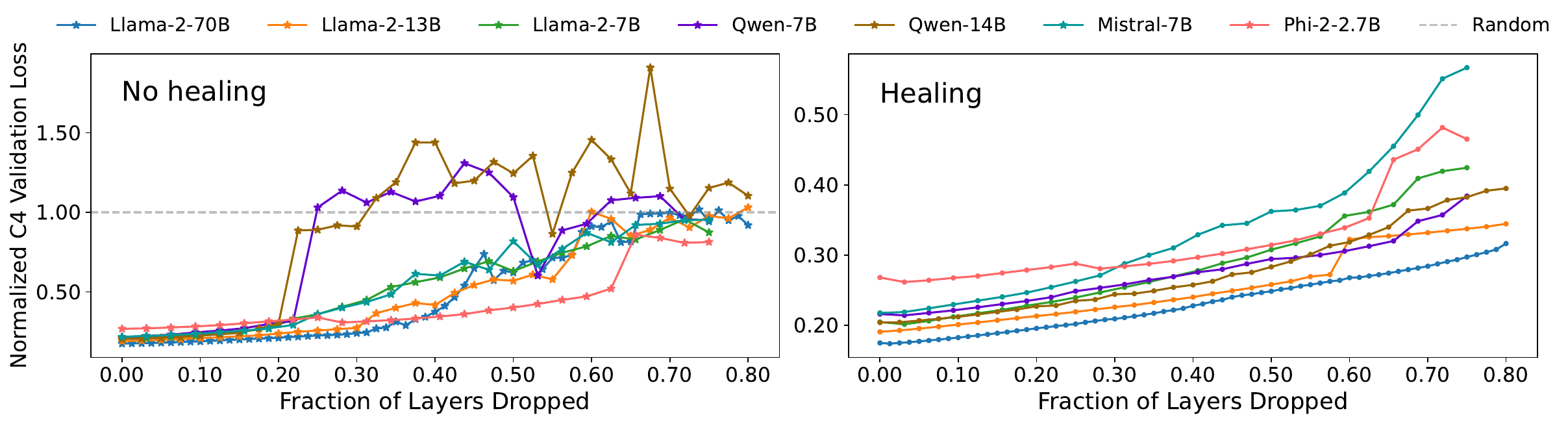}}
\caption{Normalized C4 validation loss vs. fraction of layers dropped
before healing (\emph{left}) and after healing (\emph{right});
each curve is normalized by the cross-entropy loss of sampling uniformly from the model's vocabulary.
For the experiments before healing, the loss for each model transitions to random guessing (gray dashed line) at approximately the same pruning fractions that the QA benchmarks transition to random guessing; after healing, there is continuity through the regions of sharp transition on QA tasks, cf. Figure~\ref{fig:main-results-pruning}. Contrasting the overall scale of both plots, it's clear that healing significantly restores the performance on next-token prediction to near-unpruned levels.
}
\label{fig:c4-validation-plots}
\end{center}
\end{figure}

In Figure~\ref{fig:c4-validation-plots} ,
we plot the normalized C4 validation loss for 
all seven of our models,
after healing (left panel) and before healing (right panel), 
as a 
function of the fraction layers removed. Without healing, we see that there is a somewhat sharp(ish) transition to random guessing for each model at 
approximately
the pruning fraction that the QA benchmark accuracies also sharply transition to random guessing, suggesting that models are hopelessly harmed at this point, cf. Figure~\ref{fig:main-results-pruning}.
Next, contrasting the scales of both plots, we see that healing significantly restores the next-token prediction ability of all the models to near-unpruned levels,
with
the loss 
increasing slowly and linearly with layer dropping.
Most strikingly -- from a scientific perspective -- is 
the post-healing
continuity 
through the pruning fractions where we previously found
sharp transitions
for 
the 
QA benchmarks:
this decoupling 
illustrates 
one way of
disconnecting (or creating a miscalibration) between performance on downstream tasks -- such as MMLU and BoolQ -- and continuous measures of performance -- such as the cross-entropy loss.
\footnote{This 
is consistent with \citet{schaeffer2023emergent} 
that
argued 
jumps in one kind of metric may not be visible in others.
}

\subsection{Angular distances between representations}
\label{subsec:activation}

Given the central role the angular distance \eqref{eq:arccos-sim} plays in our pruning strategy, 
let's take a subsection to look at these distances across our seven models. 
For this analysis, the angular distances for each model were averaged over 10k samples from the C4 validation set.

Recall from earlier Figure~\ref{fig:schematic-llama2-70b}(c): for Llama-2-70B this plotted the angular distance $d(x^{(\ell)}, x^{(\ell+n)})$ 
that compared the $\ell$-th layer to the $(\ell+n)$-th layer, across all initial indexes $\ell$ for block sizes from $n=1$ to $n=64$;
the minimum of the curves, $\ell^\star(n)$, gave the optimal block to prune for a given $n$, cf. \eqref{eq:optimal-block}.

\begin{figure}[t]
\begin{center}
\centerline{\includegraphics[width=1.0\columnwidth]{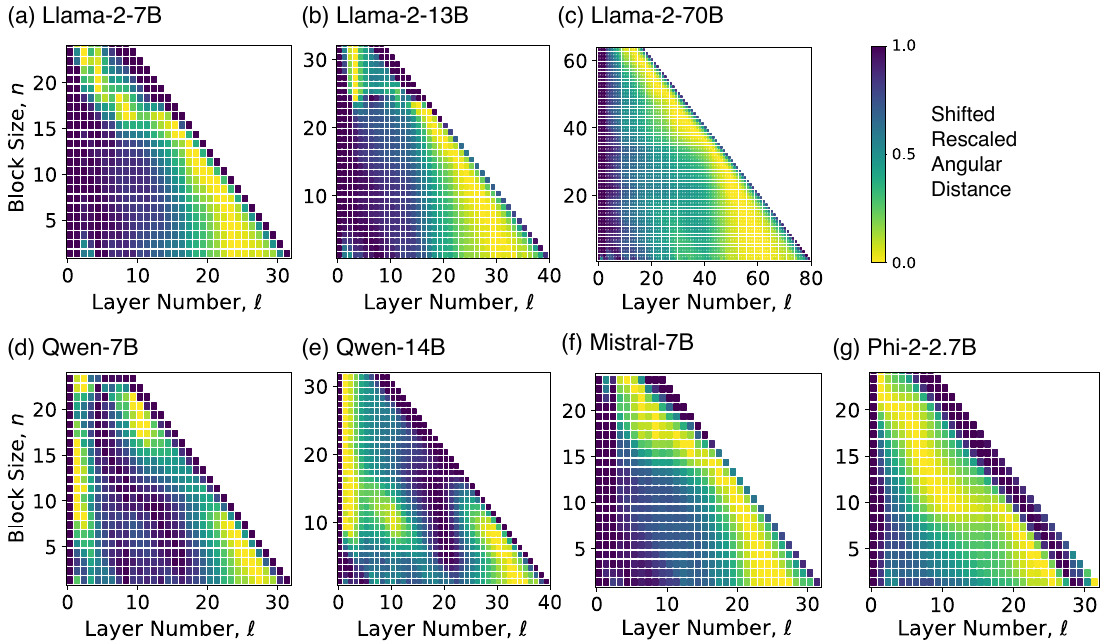}}
\caption{
Normalized angular distance \eqref{eq:arccos-sim}
from initial layer $\ell$ (x-axis) with block size $n$ (y-axis) for each of the seven models we evaluated;
the distance 
for each 
$n$
is shifted and rescaled
to span the same range, $[0,1]$ 
(yellow to purple): the optimal block to prune, $\ell^*(n)$, 
corresponds to
the deepest yellow for each row.
Across models, 
the
deeper layers tend 
to be very similar,
though 
the deepest blocks that include the final layer 
(squares along the outer diagonal)
are 
(near-)maximally dissimilar.
}
\label{fig:angular-distance-color}
\end{center}
\end{figure}

A more compact way
to display this same data
is shown in the heat maps of Figure~\ref{fig:angular-distance-color}: each square is colored to depict the row-normalized angular distance 
between layer $\ell$ and $\ell+n$ 
across all possible $\ell$, 
and $n$ up to very large fractions of the total number of layers;
the optimal layer to prune for a given block size, $\ell^*(n)$, corresponds to the minimal distance in each row.

Across models, we 
make two generalizations:
\emph{(i)} the smallest distances
are found across the deeper blocks, 
meaning 
deeper layers are typically quite similar to each other and can be more easily dropped;
\emph{(ii)} the distances across the deepest blocks -- the blocks that include the last layer
-- take either maximal or nearly-maximal values,
meaning one should never drop the final layer.
While broadly true, there
are a few exceptions. 
For some models, e.g. Phi-2-2.7B, or for the largest blocks in some models, e.g. Llama-2-7B, final \emph{few} layers seem important. 
As previously noted, the Qwen family is somewhat unusual: here we see 
that
there are a few 
odd ``islands'' of high similarity for shallow blocks; 
this
likely
explains the shorter region of robust performance in Figure~\ref{fig:main-results-pruning}.

\subsection{A simpler pruning strategy}
\label{subsec:other_patterns}

Inspired by 
our recent conclusions,
we experiment with a very simple heuristic pruning strategy: \emph{(1)} if pruning $n$ layers from an $L$-layer model, drop layers $(L-n)$ to $(L-1)$
so as to remove
the deepest block that excludes the final layer; 
then \emph{(2)} heal with a small amount of finetuning as before.
Compared with our principal similarity-informed pruning strategy, this simpler heuristic algorithm has the advantage of 
never
requiring 
practitioners
to 
load onto a GPU or inference the unpruned model. 
It also provides a meaningful ablation of the importance 
of optimizing the
block to prune. %

\begin{figure}[t]
\begin{center}
\centerline{\includegraphics[width=1.0\columnwidth]{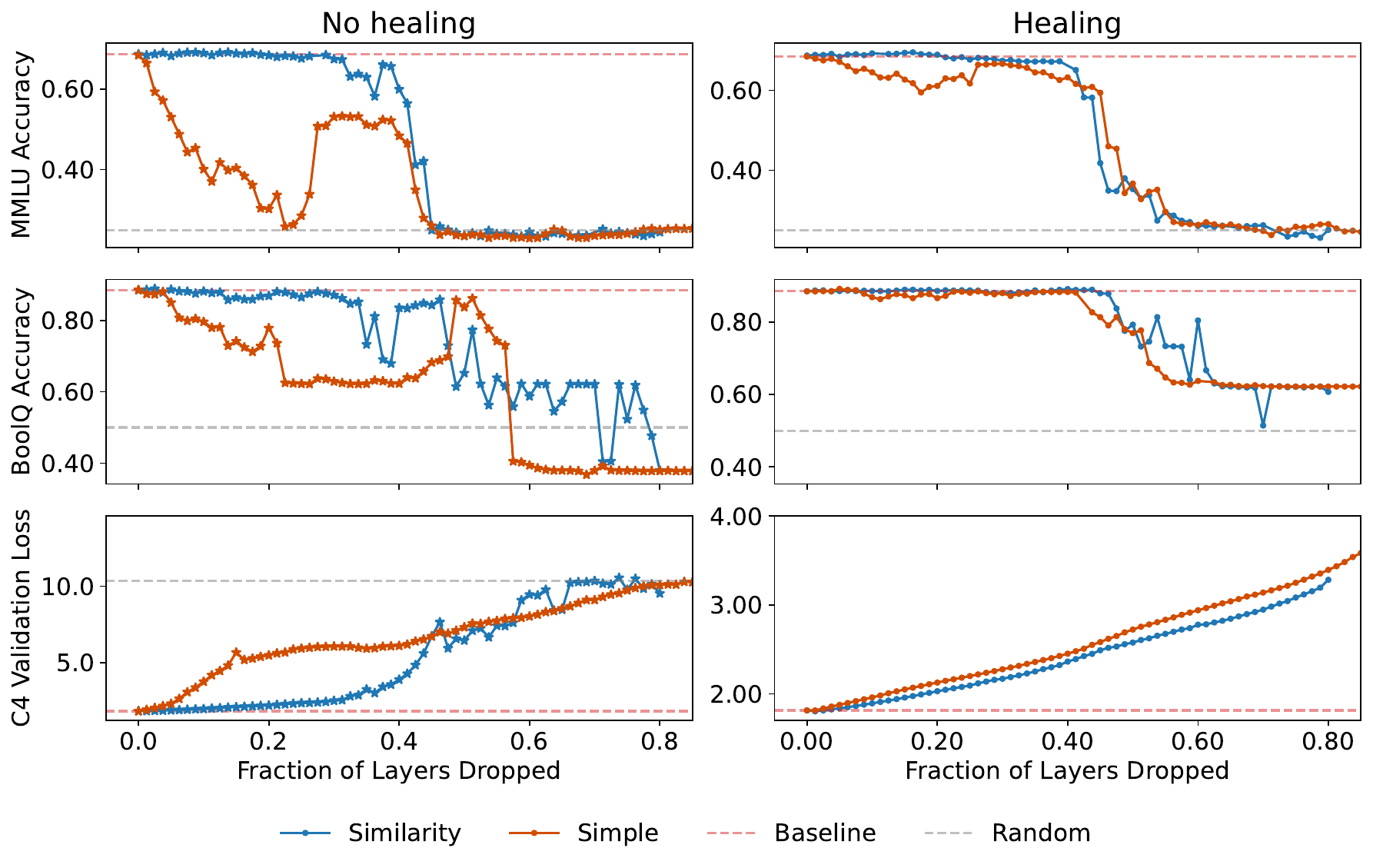}}
\caption{Evaluation of Llama-2-70B 
with
the 
simple pruning heuristic (solid red line), 
shown along with 
scores 
for
the similarity-informed pruning strategy (solid blue line), 
scores of the unpruned Llama-2-70B (red dashed line), 
and
scores for
randomly guessing (gray dashed line).
(\emph{Left:} before healing, \emph{Right:} after healing; \emph{Top:} MMLU, \emph{Middle:} BoolQ, \emph{Bottom:} C4 Validation Loss.)
Without healing, the simple heuristic performs poorly across all evals; with healing, the scores of both methods are quite similar.
}
\label{fig:simple-vs-similarity}
\end{center}
\end{figure}

In Figure~\ref{fig:simple-vs-similarity}, we contrast our two pruning strategies, both before healing (left panels) and after healing (right panels), for the QA benchmarks (MMLU/BoolQ, top/middle panels) and the autoregressive loss (C4 validation, bottom panels). On the one hand, 
the simple heuristic performs quite poorly without healing the damage incurred by pruning: accuracy on the QA benchmarks decays rapidly to (near-) random with increased pruning fraction, and the loss begins to increase very rapidly even with small amounts of pruning. 
On the other hand, 
the results for the two pruning strategies across evaluations are quite comparable after healing: for the QA benchmarks, the similarity-informed algorithm slightly better preserves the accuracy before the phase transition, 
though the simple algorithm perhaps pushes the phase transition 
to slightly 
greater pruning factions; 
and 
for the loss, 
the curves nearly lie on top of each other,
though the similarity-informed strategy 
does marginally outperform
for all amounts of pruning.
These experiments are strong evidence that the purpose of 
post-pruning finetuning is 
the 
healing 
of damage
at the pruning interface 
and not
the
acquisition of additional knowledge.

\section{Discussion and Future Directions}
\label{sec:conclusion}

At the end of this work, many readers are puzzled by the following: are the deeper layers entirely useless?
So far,
we've provided evidence that the elimination of the deeper layers does not affect performance on QA tasks like MMLU (Figure~\ref{fig:main-results-pruning}), 
while at the same time have shown that their removal does 
disrupt
the next-token predictions of the underlying model (Figure~\ref{fig:c4-validation-plots}). Since perplexity often correlates with performance on downstream tasks, 
which are the tasks that are hurt by layer pruning?

\begin{figure}[t]
\begin{center}
\centerline{\includegraphics[width=1.0\columnwidth]{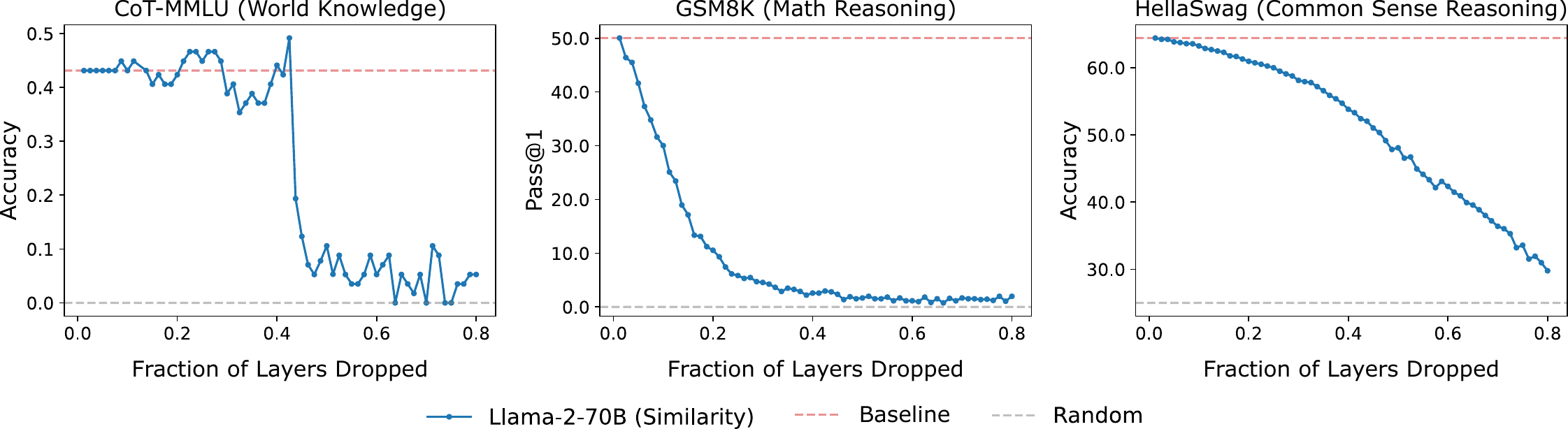}}
\caption{Evaluation of Llama-2 70B 
with the similarity-informed pruning strategy
across different evaluation tasks.  
(\emph{Left:} Chain-of-Thought MMLU (CoT-MMLU), \emph{Center:} GSM8K, \emph{Right:} HellaSwag.)
We see that GSM8K and HellaSwag show immediate degradation of performance with any level of pruning, while CoT-MMLU behaves qualitatively similarly to MMLU without CoT; this suggests that the deeper layers are likely necessary for reasoning tasks.
}
\label{fig:diff-evals-70b}
\end{center}
\end{figure}

Here are two hypotheses consistent with the fact that the model's perplexity is disturbed proportionally to pruning fraction:
\begin{itemize}
\item[\emph{(i)}] The deeper layers are not essential for storing knowledge, but are useful for more complicated computations, such as those that involve reasoning.
\item[\emph{(ii)}] The deeper layers are necessary when the model has to generate many tokens before answering a question, such as when it produces a chain-of-thought (CoT).
\end{itemize}
We test these hypotheses by evaluating our layer-pruned models on tasks that involve CoTs or reasoning. For the former, we'll look at Chain-of-Thought MMLU (CoT-MMLU); for the latter, we'll look at GSM8K \citep{cobbe2021training}, a grade-school math benchmark, and  HellaSwag \citep{zellers2019hellaswag}, a multiple choice common-sense reasoning benchmark.\footnote{Here are the details for how we performed these three evaluations:
\begin{itemize}
\item For \textbf{CoT-MMLU}, we followed the \texttt{flan\_cot\_fewshot} evaluation in EleutherAI \citep{eval-harness}, in which models produce a chain of thought before generating their answer. Note that 
the 
accuracy at $0\%$ pruning fraction 
for MMLU without CoT  
is much
better 
than the analogous accuracy at $0\%$ pruning fraction for CoT-MMLU 
($\sim 69\%$ vs. $\sim 43\%$, respectively; cf. Figures~\ref{fig:main-results-pruning}~and~\ref{fig:diff-evals-70b}),
consistent with some previous work (e.g., see Table 16 of \citet{chung2024scaling}).
\item For \textbf{GSM8K}, we used the \texttt{gsm8k\_cot} evaluation in EleutherAI \citep{eval-harness} and measured \texttt{pass@1};
for each problem we extracted an answer from a single generation (with CoT) and checked for correctness against the ground-truth answer. 

\item For \textbf{HellaSwag}, we used the \texttt{hellaswag} evaluation in EleutherAI \citep{eval-harness}. Note that HellaSwag is a multiple-choice benchmark, so random performance is 25\%.
\end{itemize}
}

In Figure~\ref{fig:diff-evals-70b}, 
we plot the performance of Llama-2 70B pruned with the similarity-informed pruning strategy across  CoT-MMLU (left), GSM8K (center), and HellaSwag (right): 
on the one hand, both GSM8K and HellaSwag, our two reasoning tasks, exhibit immediate degradation in performance with any amount of pruning, correlating with a similar decrease in the perplexity evals (Figure~\ref{fig:c4-validation-plots});
on the other hand, 
CoT-MMLU 
shows a relatively flat region of robust performance with pruning, analogous to our previous results on QA benchmarks (e.g. Figure~\ref{fig:main-results-pruning}). 
This is some initial evidence for hypothesis \emph{(i)} over hypothesis \emph{(ii)}:
the deeper layers may be useful for higher-level reasoning tasks,
while less important for knowledge intensive QA tasks; 
moreover,
perplexity errors due to pruning do not compound to hurt QA evals
when the model is required to generate many tokens.

Now at the conclusion of the work, we are left with the following questions:

\begin{itemize}
\item What are
better layer-pruning strategies? 
What are
better approaches to healing?\footnote{At the cost of introducing another hyperparameter and requiring both pruned and unpruned models to fit in memory during finetuning, one natural way to improve healing
is 
by adding
an auxiliary student-teacher loss that explicitly addresses the pruning mismatch \eqref{eq:mismatch-n}, such as
\begin{equation}
\mathcal{L}_{\text{aux}} \sim \left(x^{(\ell^*\! + n)}(\theta_0) - x^{(\ell^*)}(\theta) \right)^2 \, ,
\end{equation}
where $\theta_0$ are the frozen parameters of the unpruned model, and $\theta$ are the parameters of the pruned model to be healed; thus, $x^{(\ell^*\! + n)}(\theta_0)$ is the input to 
the $(\ell^*\! + n)$-th layer
in the unpruned model, 
$x^{(\ell^*)}(\theta)$ is the input to that same layer after pruning,
and $\mathcal{L}_{\text{aux}}$ minimizes their mismatch. %
 We thank Sho Yaida for 
 this observation.
}
\item 
Why does healing 
eliminate the phase transition in the loss but not in the QA accuracies?
\item With more comprehensive evals, will accuracy on different tasks degrade at different depths?
\item Relatedly, is knowledge generally
stored in 
shallow or middle layers,
or is it delocalized?
\item Can we devise a pruning strategy that is robust for reasoning tasks? 
\item Do pretraining details 
affect 
the
ability to prune, e.g., are scaling-law
over-trained or distilled models more difficult to prune?
\item How can we enable 
LLMs
to more effectively use the parameters in their deepest layers?
\end{itemize}
Some of these questions would benefit from studying both layer similarity and pruning across different pretraining checkpoints; for instance, at what point does the sharp phase transition and critical depth in the QA accuracies emerge, 
and does more training lead to better use of the prunable parameters? 
Others
suggest explorations with different pretraining architectures and objectives, e.g. in order
better make use of the deeper layers (for example, one can imagine applying layer dropout \citep{fan2019reducing} or early exit during pre-training \citep{elhoushi2024layer} to induce equal usage of layers).
With more comprehensive evaluations, if different kinds of QA tasks degrade at very different depths, then 
this
might indicate that
the knowledge required to complete those tasks
is
stored 
across different layers.\footnote{Alternatively, one could measure $d(x^{(\ell)},x^{(\ell+n)})$ or find
$\ell^*(n)$ as a function of 
different eval datasets. %
}
It would be very interesting to use pruning to systematically study these kind of interpretability questions.

\begin{ack}
We thank Aaron Schwartz for his initial collaboration,
Aaditya Singh 
and
Sho Yaida for discussions,
and
Aaditya Singh 
for comments on the draft.
We would also like to acknowledge the 2023 NeurIPS Large Language Model Efficiency Challenge 
for initializing us 
for
work on this project.
A.G. is supported by the NSF CAREER grant DMR-2045181, the Sloan Foundation, and
by the Laboratory for Physical Sciences through the Condensed Matter Theory Center.
D.R. acknowledges support from the National Science Foundation under Cooperative Agreement PHY-2019786 (the NSF AI Institute for Artificial Intelligence
and Fundamental Interactions, \texttt{http://iaifi.org/}) and appreciates both the sanction and support of Sequoia Capital. 
This paper has been 
brought to you residually by the letters $G$, $P$, and $U$, after summing over many layers.

\end{ack}

\bibliography{iclr2025_conference}

\begin{thebibliography}{81}
\providecommand{\natexlab}[1]{#1}
\providecommand{\url}[1]{\texttt{#1}}
\expandafter\ifx\csname urlstyle\endcsname\relax
  \providecommand{\doi}[1]{doi: #1}\else
  \providecommand{\doi}{doi: \begingroup \urlstyle{rm}\Url}\fi

\bibitem[Touvron et~al.(2023)Touvron, Martin, Stone, Albert, Almahairi, Babaei, Bashlykov, Batra, Bhargava, Bhosale, et~al.]{touvron2023llama2}
Hugo Touvron, Louis Martin, Kevin Stone, Peter Albert, Amjad Almahairi, Yasmine Babaei, Nikolay Bashlykov, Soumya Batra, Prajjwal Bhargava, Shruti Bhosale, et~al.
\newblock Llama 2: Open foundation and fine-tuned chat models.
\newblock \emph{arXiv preprint arXiv:2307.09288}, 2023.

\bibitem[nostalgebraist(2020)]{logit_lens_2020}
nostalgebraist.
\newblock interpreting gpt: the logit lens.
\newblock \url{https://www.lesswrong.com/posts/AcKRB8wDpdaN6v6ru/interpreting-gpt-the-logit-lens}, 2020.

\bibitem[Belrose et~al.(2023)Belrose, Furman, Smith, Halawi, Ostrovsky, McKinney, Biderman, and Steinhardt]{belrose2023eliciting}
Nora Belrose, Zach Furman, Logan Smith, Danny Halawi, Igor Ostrovsky, Lev McKinney, Stella Biderman, and Jacob Steinhardt.
\newblock Eliciting latent predictions from transformers with the tuned lens.
\newblock \emph{arXiv preprint arXiv:2303.08112}, 2023.

\bibitem[Chen et~al.(2018)Chen, Rubanova, Bettencourt, and Duvenaud]{chen2018neural}
Ricky~TQ Chen, Yulia Rubanova, Jesse Bettencourt, and David~K Duvenaud.
\newblock Neural ordinary differential equations.
\newblock \emph{Advances in neural information processing systems}, 31, 2018.

\bibitem[Yang et~al.(2023)Yang, Yu, Zhu, and Hayou]{yang2023tensor}
Greg Yang, Dingli Yu, Chen Zhu, and Soufiane Hayou.
\newblock Tensor programs vi: Feature learning in infinite-depth neural networks.
\newblock \emph{arXiv preprint arXiv:2310.02244}, 2023.

\bibitem[LeCun et~al.(1989)LeCun, Denker, and Solla]{NIPS1989_6c9882bb}
Yann LeCun, John Denker, and Sara Solla.
\newblock Optimal brain damage.
\newblock In D.~Touretzky, editor, \emph{Advances in Neural Information Processing Systems}, volume~2. Morgan-Kaufmann, 1989.

\bibitem[Hassibi and Stork(1992)]{NIPS1992_303ed4c6}
Babak Hassibi and David Stork.
\newblock Second order derivatives for network pruning: Optimal brain surgeon.
\newblock In S.~Hanson, J.~Cowan, and C.~Giles, editors, \emph{Advances in Neural Information Processing Systems}, volume~5. Morgan-Kaufmann, 1992.

\bibitem[Han et~al.(2015)Han, Pool, Tran, and Dally]{han2015learning}
Song Han, Jeff Pool, John Tran, and William Dally.
\newblock Learning both weights and connections for efficient neural network.
\newblock \emph{Advances in neural information processing systems}, 28, 2015.

\bibitem[Chen et~al.(2015)Chen, Wilson, Tyree, Weinberger, and Chen]{chen2015compressing}
Wenlin Chen, James Wilson, Stephen Tyree, Kilian Weinberger, and Yixin Chen.
\newblock Compressing neural networks with the hashing trick.
\newblock In \emph{International conference on machine learning}, pages 2285--2294. PMLR, 2015.

\bibitem[Srinivas and Babu(2015)]{srinivas2015data}
Suraj Srinivas and R~Venkatesh Babu.
\newblock Data-free parameter pruning for deep neural networks.
\newblock \emph{arXiv preprint arXiv:1507.06149}, 2015.

\bibitem[Li et~al.(2016)Li, Kadav, Durdanovic, Samet, and Graf]{li2016pruning}
Hao Li, Asim Kadav, Igor Durdanovic, Hanan Samet, and Hans~Peter Graf.
\newblock Pruning filters for efficient convnets.
\newblock \emph{arXiv preprint arXiv:1608.08710}, 2016.

\bibitem[Wen et~al.(2016)Wen, Wu, Wang, Chen, and Li]{wen2016learning}
Wei Wen, Chunpeng Wu, Yandan Wang, Yiran Chen, and Hai Li.
\newblock Learning structured sparsity in deep neural networks.
\newblock \emph{Advances in neural information processing systems}, 29, 2016.

\bibitem[Hu et~al.(2016)Hu, Peng, Tai, and Tang]{hu2016network}
Hengyuan Hu, Rui Peng, Yu-Wing Tai, and Chi-Keung Tang.
\newblock Network trimming: A data-driven neuron pruning approach towards efficient deep architectures.
\newblock \emph{arXiv preprint arXiv:1607.03250}, 2016.

\bibitem[He et~al.(2017)He, Zhang, and Sun]{he2017channel}
Yihui He, Xiangyu Zhang, and Jian Sun.
\newblock Channel pruning for accelerating very deep neural networks.
\newblock In \emph{Proceedings of the IEEE international conference on computer vision}, pages 1389--1397, 2017.

\bibitem[Huang et~al.(2018)Huang, Liu, Van~der Maaten, and Weinberger]{huang2018condensenet}
Gao Huang, Shichen Liu, Laurens Van~der Maaten, and Kilian~Q Weinberger.
\newblock Condensenet: An efficient densenet using learned group convolutions.
\newblock In \emph{Proceedings of the IEEE conference on computer vision and pattern recognition}, pages 2752--2761, 2018.

\bibitem[Murray and Chiang(2015)]{murray2015auto}
Kenton Murray and David Chiang.
\newblock Auto-sizing neural networks: With applications to n-gram language models.
\newblock \emph{arXiv preprint arXiv:1508.05051}, 2015.

\bibitem[See et~al.(2016)See, Luong, and Manning]{see2016compression}
Abigail See, Minh-Thang Luong, and Christopher~D Manning.
\newblock Compression of neural machine translation models via pruning.
\newblock \emph{arXiv preprint arXiv:1606.09274}, 2016.

\bibitem[Kim and Rush(2016)]{kim2016sequence}
Yoon Kim and Alexander~M Rush.
\newblock Sequence-level knowledge distillation.
\newblock \emph{arXiv preprint arXiv:1606.07947}, 2016.

\bibitem[Voita et~al.(2019)Voita, Talbot, Moiseev, Sennrich, and Titov]{voita2019analyzing}
Elena Voita, David Talbot, Fedor Moiseev, Rico Sennrich, and Ivan Titov.
\newblock Analyzing multi-head self-attention: Specialized heads do the heavy lifting, the rest can be pruned.
\newblock \emph{arXiv preprint arXiv:1905.09418}, 2019.

\bibitem[Michel et~al.(2019)Michel, Levy, and Neubig]{michel2019sixteen}
Paul Michel, Omer Levy, and Graham Neubig.
\newblock Are sixteen heads really better than one?
\newblock \emph{Advances in neural information processing systems}, 32, 2019.

\bibitem[Kim and Awadalla(2020)]{kim2020fastformers}
Young~Jin Kim and Hany~Hassan Awadalla.
\newblock Fastformers: Highly efficient transformer models for natural language understanding.
\newblock \emph{arXiv preprint arXiv:2010.13382}, 2020.

\bibitem[Fan et~al.(2019)Fan, Grave, and Joulin]{fan2019reducing}
Angela Fan, Edouard Grave, and Armand Joulin.
\newblock Reducing transformer depth on demand with structured dropout.
\newblock \emph{arXiv preprint arXiv:1909.11556}, 2019.

\bibitem[Zhang and He(2020)]{zhang2020accelerating}
Minjia Zhang and Yuxiong He.
\newblock Accelerating training of transformer-based language models with progressive layer dropping.
\newblock \emph{Advances in Neural Information Processing Systems}, 33:\penalty0 14011--14023, 2020.

\bibitem[Fan et~al.(2021)Fan, Li, Ao, Wu, Meng, and Sun]{fan2021layer}
Chun Fan, Jiwei Li, Xiang Ao, Fei Wu, Yuxian Meng, and Xiaofei Sun.
\newblock Layer-wise model pruning based on mutual information.
\newblock \emph{arXiv preprint arXiv:2108.12594}, 2021.

\bibitem[Jha et~al.(2023)Jha, Groeneveld, Strubell, and Beltagy]{jha2023large}
Ananya~Harsh Jha, Dirk Groeneveld, Emma Strubell, and Iz~Beltagy.
\newblock Large language model distillation doesn't need a teacher.
\newblock \emph{arXiv preprint arXiv:2305.14864}, 2023.

\bibitem[Sajjad et~al.(2023)Sajjad, Dalvi, Durrani, and Nakov]{sajjad2023effect}
Hassan Sajjad, Fahim Dalvi, Nadir Durrani, and Preslav Nakov.
\newblock On the effect of dropping layers of pre-trained transformer models.
\newblock \emph{Computer Speech \& Language}, 77:\penalty0 101429, 2023.

\bibitem[Liu et~al.(2023{\natexlab{a}})Liu, Peng, and Lee]{liu2023comflp}
Wei Liu, Zhiyuan Peng, and Tan Lee.
\newblock Comflp: Correlation measure based fast search on asr layer pruning.
\newblock \emph{arXiv preprint arXiv:2309.11768}, 2023{\natexlab{a}}.

\bibitem[Hou et~al.(2020)Hou, Huang, Shang, Jiang, Chen, and Liu]{hou2020dynabert}
Lu~Hou, Zhiqi Huang, Lifeng Shang, Xin Jiang, Xiao Chen, and Qun Liu.
\newblock Dynabert: Dynamic bert with adaptive width and depth.
\newblock \emph{Advances in Neural Information Processing Systems}, 33:\penalty0 9782--9793, 2020.

\bibitem[Sharma et~al.(2023)Sharma, Ash, and Misra]{sharma2023truth}
Pratyusha Sharma, Jordan~T Ash, and Dipendra Misra.
\newblock The truth is in there: Improving reasoning in language models with layer-selective rank reduction.
\newblock \emph{arXiv preprint arXiv:2312.13558}, 2023.

\bibitem[Ashkboos et~al.(2024)Ashkboos, Croci, Gennari~do Nascimento, Hoefler, and Hensman]{ashkboos2024sliceGPT}
Saleh Ashkboos, Maximilian~L. Croci, Marcelo Gennari~do Nascimento, Torsten Hoefler, and James Hensman.
\newblock Slicegpt: Compress large language models by deleting rows and columns.
\newblock \emph{arXiv preprint arXiv:2401.15024}, 2024.

\bibitem[Xia et~al.(2022)Xia, Zhong, and Chen]{xia2022structured}
Mengzhou Xia, Zexuan Zhong, and Danqi Chen.
\newblock Structured pruning learns compact and accurate models.
\newblock \emph{arXiv preprint arXiv:2204.00408}, 2022.

\bibitem[Lagunas et~al.(2021)Lagunas, Charlaix, Sanh, and Rush]{lagunas2021block}
Fran{\c{c}}ois Lagunas, Ella Charlaix, Victor Sanh, and Alexander~M Rush.
\newblock Block pruning for faster transformers.
\newblock \emph{arXiv preprint arXiv:2109.04838}, 2021.

\bibitem[Men et~al.(2024)Men, Xu, Zhang, Wang, Lin, Lu, Han, and Chen]{men2024shortgpt}
Xin Men, Mingyu Xu, Qingyu Zhang, Bingning Wang, Hongyu Lin, Yaojie Lu, Xianpei Han, and Weipeng Chen.
\newblock Shortgpt: Layers in large language models are more redundant than you expect.
\newblock \emph{arXiv preprint arXiv:2403.03853}, 2024.

\bibitem[Bai et~al.(2023)Bai, Bai, Chu, Cui, Dang, Deng, Fan, Ge, Han, Huang, et~al.]{bai2023qwen}
Jinze Bai, Shuai Bai, Yunfei Chu, Zeyu Cui, Kai Dang, Xiaodong Deng, Yang Fan, Wenbin Ge, Yu~Han, Fei Huang, et~al.
\newblock Qwen technical report.
\newblock \emph{arXiv preprint arXiv:2309.16609}, 2023.

\bibitem[Jiang et~al.(2023{\natexlab{a}})Jiang, Sablayrolles, Mensch, Bamford, Chaplot, Casas, Bressand, Lengyel, Lample, Saulnier, et~al.]{jiang2023mistral}
Albert~Q Jiang, Alexandre Sablayrolles, Arthur Mensch, Chris Bamford, Devendra~Singh Chaplot, Diego de~las Casas, Florian Bressand, Gianna Lengyel, Guillaume Lample, Lucile Saulnier, et~al.
\newblock Mistral 7b.
\newblock \emph{arXiv preprint arXiv:2310.06825}, 2023{\natexlab{a}}.

\bibitem[Javaheripi and Bubeck(2023)]{phi2}
Mojan Javaheripi and S{\'e}bastien Bubeck.
\newblock Phi-2: The surprising power of small language models, Dec 2023.

\bibitem[Dettmers et~al.(2023)Dettmers, Pagnoni, Holtzman, and Zettlemoyer]{dettmers2023qlora}
Tim Dettmers, Artidoro Pagnoni, Ari Holtzman, and Luke Zettlemoyer.
\newblock Qlora: Efficient finetuning of quantized llms.
\newblock \emph{arXiv preprint arXiv:2305.14314}, 2023.

\bibitem[Raffel et~al.(2020)Raffel, Shazeer, Roberts, Lee, Narang, Matena, Zhou, Li, and Liu]{raffel2020exploring}
Colin Raffel, Noam Shazeer, Adam Roberts, Katherine Lee, Sharan Narang, Michael Matena, Yanqi Zhou, Wei Li, and Peter~J Liu.
\newblock Exploring the limits of transfer learning with a unified text-to-text transformer.
\newblock \emph{The Journal of Machine Learning Research}, 21\penalty0 (1):\penalty0 5485--5551, 2020.

\bibitem[Hendrycks et~al.(2020)Hendrycks, Burns, Basart, Zou, Mazeika, Song, and Steinhardt]{hendrycks2020measuring}
Dan Hendrycks, Collin Burns, Steven Basart, Andy Zou, Mantas Mazeika, Dawn Song, and Jacob Steinhardt.
\newblock Measuring massive multitask language understanding.
\newblock \emph{arXiv preprint arXiv:2009.03300}, 2020.

\bibitem[Clark et~al.(2019)Clark, Lee, Chang, Kwiatkowski, Collins, and Toutanova]{clark2019boolq}
Christopher Clark, Kenton Lee, Ming-Wei Chang, Tom Kwiatkowski, Michael Collins, and Kristina Toutanova.
\newblock Boolq: Exploring the surprising difficulty of natural yes/no questions.
\newblock \emph{arXiv preprint arXiv:1905.10044}, 2019.

\bibitem[Schaeffer et~al.(2023)Schaeffer, Miranda, and Koyejo]{schaeffer2023emergent}
Rylan Schaeffer, Brando Miranda, and Sanmi Koyejo.
\newblock Are emergent abilities of large language models a mirage?
\newblock \emph{arXiv preprint arXiv:2304.15004}, 2023.

\bibitem[Cobbe et~al.(2021)Cobbe, Kosaraju, Bavarian, Chen, Jun, Kaiser, Plappert, Tworek, Hilton, Nakano, et~al.]{cobbe2021training}
Karl Cobbe, Vineet Kosaraju, Mohammad Bavarian, Mark Chen, Heewoo Jun, Lukasz Kaiser, Matthias Plappert, Jerry Tworek, Jacob Hilton, Reiichiro Nakano, et~al.
\newblock Training verifiers to solve math word problems.
\newblock \emph{arXiv preprint arXiv:2110.14168}, 2021.

\bibitem[Zellers et~al.(2019)Zellers, Holtzman, Bisk, Farhadi, and Choi]{zellers2019hellaswag}
Rowan Zellers, Ari Holtzman, Yonatan Bisk, Ali Farhadi, and Yejin Choi.
\newblock Hellaswag: Can a machine really finish your sentence?
\newblock \emph{arXiv preprint arXiv:1905.07830}, 2019.

\bibitem[Gao et~al.(2023)Gao, Tow, Abbasi, Biderman, Black, DiPofi, Foster, Golding, Hsu, Le~Noac'h, Li, McDonell, Muennighoff, Ociepa, Phang, Reynolds, Schoelkopf, Skowron, Sutawika, Tang, Thite, Wang, Wang, and Zou]{eval-harness}
Leo Gao, Jonathan Tow, Baber Abbasi, Stella Biderman, Sid Black, Anthony DiPofi, Charles Foster, Laurence Golding, Jeffrey Hsu, Alain Le~Noac'h, Haonan Li, Kyle McDonell, Niklas Muennighoff, Chris Ociepa, Jason Phang, Laria Reynolds, Hailey Schoelkopf, Aviya Skowron, Lintang Sutawika, Eric Tang, Anish Thite, Ben Wang, Kevin Wang, and Andy Zou.
\newblock A framework for few-shot language model evaluation, 12 2023.
\newblock URL \url{https://zenodo.org/records/10256836}.

\bibitem[Chung et~al.(2024)Chung, Hou, Longpre, Zoph, Tay, Fedus, Li, Wang, Dehghani, Brahma, et~al.]{chung2024scaling}
Hyung~Won Chung, Le~Hou, Shayne Longpre, Barret Zoph, Yi~Tay, William Fedus, Yunxuan Li, Xuezhi Wang, Mostafa Dehghani, Siddhartha Brahma, et~al.
\newblock Scaling instruction-finetuned language models.
\newblock \emph{Journal of Machine Learning Research}, 25\penalty0 (70):\penalty0 1--53, 2024.

\bibitem[Elhoushi et~al.(2024)Elhoushi, Shrivastava, Liskovich, Hosmer, Wasti, Lai, Mahmoud, Acun, Agarwal, Roman, et~al.]{elhoushi2024layer}
Mostafa Elhoushi, Akshat Shrivastava, Diana Liskovich, Basil Hosmer, Bram Wasti, Liangzhen Lai, Anas Mahmoud, Bilge Acun, Saurabh Agarwal, Ahmed Roman, et~al.
\newblock Layer skip: Enabling early exit inference and self-speculative decoding.
\newblock \emph{arXiv preprint arXiv:2404.16710}, 2024.

\bibitem[Vaswani et~al.(2017)Vaswani, Shazeer, Parmar, Uszkoreit, Jones, Gomez, Kaiser, and Polosukhin]{vaswani2017attention}
Ashish Vaswani, Noam Shazeer, Niki Parmar, Jakob Uszkoreit, Llion Jones, Aidan~N Gomez, {\L}ukasz Kaiser, and Illia Polosukhin.
\newblock Attention is all you need.
\newblock \emph{Advances in neural information processing systems}, 30, 2017.

\bibitem[Devlin et~al.(2018)Devlin, Chang, Lee, and Toutanova]{devlin2018bert}
Jacob Devlin, Ming-Wei Chang, Kenton Lee, and Kristina Toutanova.
\newblock Bert: Pre-training of deep bidirectional transformers for language understanding.
\newblock \emph{arXiv preprint arXiv:1810.04805}, 2018.

\bibitem[Radford et~al.(2019)Radford, Wu, Child, Luan, Amodei, and Sutskever]{Radford2019LanguageMA}
Alec Radford, Jeff Wu, Rewon Child, David Luan, Dario Amodei, and Ilya Sutskever.
\newblock Language models are unsupervised multitask learners.
\newblock 2019.
\newblock URL \url{https://cdn.openai.com/better-language-models/language_models_are_unsupervised_multitask_learners.pdf}.

\bibitem[Zhong et~al.(2023)Zhong, Ding, Liu, Du, and Tao]{zhong2023can}
Qihuang Zhong, Liang Ding, Juhua Liu, Bo~Du, and Dacheng Tao.
\newblock Can chatgpt understand too? a comparative study on chatgpt and fine-tuned bert.
\newblock \emph{arXiv preprint arXiv:2302.10198}, 2023.

\bibitem[Ethayarajh(2019)]{ethayarajh2019contextual}
Kawin Ethayarajh.
\newblock How contextual are contextualized word representations? comparing the geometry of bert, elmo, and gpt-2 embeddings.
\newblock \emph{arXiv preprint arXiv:1909.00512}, 2019.

\bibitem[Baevski et~al.(2020)Baevski, Zhou, Mohamed, and Auli]{baevski2020wav2vec}
Alexei Baevski, Yuhao Zhou, Abdelrahman Mohamed, and Michael Auli.
\newblock wav2vec 2.0: A framework for self-supervised learning of speech representations.
\newblock \emph{Advances in neural information processing systems}, 33:\penalty0 12449--12460, 2020.

\bibitem[Hinton et~al.(2015)Hinton, Vinyals, and Dean]{hinton2015distilling}
Geoffrey Hinton, Oriol Vinyals, and Jeff Dean.
\newblock Distilling the knowledge in a neural network.
\newblock \emph{arXiv preprint arXiv:1503.02531}, 2015.

\bibitem[Gu et~al.(2023)Gu, Dong, Wei, and Huang]{gu2023knowledge}
Yuxian Gu, Li~Dong, Furu Wei, and Minlie Huang.
\newblock Knowledge distillation of large language models.
\newblock \emph{arXiv preprint arXiv:2306.08543}, 2023.

\bibitem[Jiao et~al.(2019)Jiao, Yin, Shang, Jiang, Chen, Li, Wang, and Liu]{jiao2019tinybert}
Xiaoqi Jiao, Yichun Yin, Lifeng Shang, Xin Jiang, Xiao Chen, Linlin Li, Fang Wang, and Qun Liu.
\newblock Tinybert: Distilling bert for natural language understanding.
\newblock \emph{arXiv preprint arXiv:1909.10351}, 2019.

\bibitem[Wang et~al.(2021)Wang, Liu, Xu, Zhu, and Zeng]{wang2021want}
Shuohang Wang, Yang Liu, Yichong Xu, Chenguang Zhu, and Michael Zeng.
\newblock Want to reduce labeling cost? gpt-3 can help.
\newblock \emph{arXiv preprint arXiv:2108.13487}, 2021.

\bibitem[Eldan and Li(2023)]{eldan2023tinystories}
Ronen Eldan and Yuanzhi Li.
\newblock Tinystories: How small can language models be and still speak coherent english?
\newblock \emph{arXiv preprint arXiv:2305.07759}, 2023.

\bibitem[Li et~al.(2023{\natexlab{a}})Li, Bubeck, Eldan, Del~Giorno, Gunasekar, and Lee]{li2023textbooks}
Yuanzhi Li, S{\'e}bastien Bubeck, Ronen Eldan, Allie Del~Giorno, Suriya Gunasekar, and Yin~Tat Lee.
\newblock Textbooks are all you need ii: phi-1.5 technical report.
\newblock \emph{arXiv preprint arXiv:2309.05463}, 2023{\natexlab{a}}.

\bibitem[Gunasekar et~al.(2023)Gunasekar, Zhang, Aneja, Mendes, Del~Giorno, Gopi, Javaheripi, Kauffmann, de~Rosa, Saarikivi, et~al.]{gunasekar2023textbooks}
Suriya Gunasekar, Yi~Zhang, Jyoti Aneja, Caio C{\'e}sar~Teodoro Mendes, Allie Del~Giorno, Sivakanth Gopi, Mojan Javaheripi, Piero Kauffmann, Gustavo de~Rosa, Olli Saarikivi, et~al.
\newblock Textbooks are all you need.
\newblock \emph{arXiv preprint arXiv:2306.11644}, 2023.

\bibitem[Fu et~al.(2023)Fu, Peng, Ou, Sabharwal, and Khot]{fu2023specializing}
Yao Fu, Hao Peng, Litu Ou, Ashish Sabharwal, and Tushar Khot.
\newblock Specializing smaller language models towards multi-step reasoning.
\newblock \emph{arXiv preprint arXiv:2301.12726}, 2023.

\bibitem[Hsieh et~al.(2023)Hsieh, Li, Yeh, Nakhost, Fujii, Ratner, Krishna, Lee, and Pfister]{hsieh2023distilling}
Cheng-Yu Hsieh, Chun-Liang Li, Chih-Kuan Yeh, Hootan Nakhost, Yasuhisa Fujii, Alexander Ratner, Ranjay Krishna, Chen-Yu Lee, and Tomas Pfister.
\newblock Distilling step-by-step! outperforming larger language models with less training data and smaller model sizes.
\newblock \emph{arXiv preprint arXiv:2305.02301}, 2023.

\bibitem[Jiang et~al.(2023{\natexlab{b}})Jiang, Chan, Chen, and Wang]{jiang2023lion}
Yuxin Jiang, Chunkit Chan, Mingyang Chen, and Wei Wang.
\newblock Lion: Adversarial distillation of closed-source large language model.
\newblock \emph{arXiv preprint arXiv:2305.12870}, 2023{\natexlab{b}}.

\bibitem[Hu et~al.(2021)Hu, Shen, Wallis, Allen-Zhu, Li, Wang, Wang, and Chen]{hu2021lora}
Edward~J Hu, Yelong Shen, Phillip Wallis, Zeyuan Allen-Zhu, Yuanzhi Li, Shean Wang, Lu~Wang, and Weizhu Chen.
\newblock Lora: Low-rank adaptation of large language models.
\newblock \emph{arXiv preprint arXiv:2106.09685}, 2021.

\bibitem[Li et~al.(2023{\natexlab{b}})Li, Yu, Liang, He, Karampatziakis, Chen, and Zhao]{li2023loftq}
Yixiao Li, Yifan Yu, Chen Liang, Pengcheng He, Nikos Karampatziakis, Weizhu Chen, and Tuo Zhao.
\newblock Loftq: Lora-fine-tuning-aware quantization for large language models.
\newblock \emph{arXiv preprint arXiv:2310.08659}, 2023{\natexlab{b}}.

\bibitem[Zhang et~al.(2023)Zhang, Chen, Bukharin, He, Cheng, Chen, and Zhao]{zhang2023adaptive}
Qingru Zhang, Minshuo Chen, Alexander Bukharin, Pengcheng He, Yu~Cheng, Weizhu Chen, and Tuo Zhao.
\newblock Adaptive budget allocation for parameter-efficient fine-tuning.
\newblock \emph{arXiv preprint arXiv:2303.10512}, 2023.

\bibitem[Leviathan et~al.(2023)Leviathan, Kalman, and Matias]{leviathan2023fast}
Yaniv Leviathan, Matan Kalman, and Yossi Matias.
\newblock Fast inference from transformers via speculative decoding.
\newblock In \emph{International Conference on Machine Learning}, pages 19274--19286. PMLR, 2023.

\bibitem[Cai et~al.(2024)Cai, Li, Geng, Peng, Lee, Chen, and Dao]{cai2024medusa}
Tianle Cai, Yuhong Li, Zhengyang Geng, Hongwu Peng, Jason~D Lee, Deming Chen, and Tri Dao.
\newblock Medusa: Simple llm inference acceleration framework with multiple decoding heads.
\newblock \emph{arXiv preprint arXiv:2401.10774}, 2024.

\bibitem[Meng et~al.(2022)Meng, Bau, Andonian, and Belinkov]{meng2022locating}
Kevin Meng, David Bau, Alex Andonian, and Yonatan Belinkov.
\newblock Locating and editing factual associations in gpt.
\newblock \emph{Advances in Neural Information Processing Systems}, 35:\penalty0 17359--17372, 2022.

\bibitem[Dai et~al.(2021)Dai, Dong, Hao, Sui, Chang, and Wei]{dai2021knowledge}
Damai Dai, Li~Dong, Yaru Hao, Zhifang Sui, Baobao Chang, and Furu Wei.
\newblock Knowledge neurons in pretrained transformers.
\newblock \emph{arXiv preprint arXiv:2104.08696}, 2021.

\bibitem[Hase et~al.(2023)Hase, Bansal, Kim, and Ghandeharioun]{hase2023does}
Peter Hase, Mohit Bansal, Been Kim, and Asma Ghandeharioun.
\newblock Does localization inform editing? surprising differences in causality-based localization vs. knowledge editing in language models.
\newblock \emph{arXiv preprint arXiv:2301.04213}, 2023.

\bibitem[Geva et~al.(2023)Geva, Bastings, Filippova, and Globerson]{geva2023dissecting}
Mor Geva, Jasmijn Bastings, Katja Filippova, and Amir Globerson.
\newblock Dissecting recall of factual associations in auto-regressive language models.
\newblock \emph{arXiv preprint arXiv:2304.14767}, 2023.

\bibitem[Din et~al.(2023)Din, Karidi, Choshen, and Geva]{din2023jump}
Alexander~Yom Din, Taelin Karidi, Leshem Choshen, and Mor Geva.
\newblock Jump to conclusions: Short-cutting transformers with linear transformations.
\newblock \emph{arXiv preprint arXiv:2303.09435}, 2023.

\bibitem[Gurnee and Tegmark(2023)]{gurnee2023language}
Wes Gurnee and Max Tegmark.
\newblock Language models represent space and time.
\newblock \emph{arXiv preprint arXiv:2310.02207}, 2023.

\bibitem[Voita et~al.(2023)Voita, Ferrando, and Nalmpantis]{voita2023neurons}
Elena Voita, Javier Ferrando, and Christoforos Nalmpantis.
\newblock Neurons in large language models: Dead, n-gram, positional.
\newblock \emph{arXiv preprint arXiv:2309.04827}, 2023.

\bibitem[Liu et~al.(2023{\natexlab{b}})Liu, Wang, Dao, Zhou, Yuan, Song, Shrivastava, Zhang, Tian, Re, et~al.]{liu2023deja}
Zichang Liu, Jue Wang, Tri Dao, Tianyi Zhou, Binhang Yuan, Zhao Song, Anshumali Shrivastava, Ce~Zhang, Yuandong Tian, Christopher Re, et~al.
\newblock Deja vu: Contextual sparsity for efficient llms at inference time.
\newblock In \emph{International Conference on Machine Learning}, pages 22137--22176. PMLR, 2023{\natexlab{b}}.

\bibitem[Panigrahi et~al.(2023)Panigrahi, Saunshi, Zhao, and Arora]{panigrahi2023task}
Abhishek Panigrahi, Nikunj Saunshi, Haoyu Zhao, and Sanjeev Arora.
\newblock Task-specific skill localization in fine-tuned language models.
\newblock \emph{arXiv preprint arXiv:2302.06600}, 2023.

\bibitem[Wolf et~al.(2020)Wolf, Debut, Sanh, Chaumond, Delangue, Moi, Cistac, Rault, Louf, Funtowicz, Davison, Shleifer, von Platen, Ma, Jernite, Plu, Xu, Scao, Gugger, Drame, Lhoest, and Rush]{wolf_etal_2020_transformers}
Thomas Wolf, Lysandre Debut, Victor Sanh, Julien Chaumond, Clement Delangue, Anthony Moi, Pierric Cistac, Tim Rault, Rémi Louf, Morgan Funtowicz, Joe Davison, Sam Shleifer, Patrick von Platen, Clara Ma, Yacine Jernite, Julien Plu, Canwen Xu, Teven~Le Scao, Sylvain Gugger, Mariama Drame, Quentin Lhoest, and Alexander~M. Rush.
\newblock Transformers: State-of-the-art natural language processing.
\newblock In \emph{Proceedings of the 2020 Conference on Empirical Methods in Natural Language Processing: System Demonstrations}, pages 38--45, Online, October 2020. Association for Computational Linguistics.
\newblock URL \url{https://www.aclweb.org/anthology/2020.emnlp-demos.6}.

\bibitem[Raffel et~al.(2019)Raffel, Shazeer, Roberts, Lee, Narang, Matena, Zhou, Li, and Liu]{2019t5}
Colin Raffel, Noam Shazeer, Adam Roberts, Katherine Lee, Sharan Narang, Michael Matena, Yanqi Zhou, Wei Li, and Peter~J. Liu.
\newblock Exploring the limits of transfer learning with a unified text-to-text transformer.
\newblock \emph{arXiv e-prints}, 2019.

\bibitem[Mangrulkar et~al.(2022)Mangrulkar, Gugger, Debut, Belkada, Paul, and Bossan]{peft}
Sourab Mangrulkar, Sylvain Gugger, Lysandre Debut, Younes Belkada, Sayak Paul, and Benjamin Bossan.
\newblock Peft: State-of-the-art parameter-efficient fine-tuning methods.
\newblock \url{https://github.com/huggingface/peft}, 2022.

\bibitem[Lee et~al.(2023)Lee, Hunter, and Ruiz]{lee2023platypus}
Ariel~N Lee, Cole~J Hunter, and Nataniel Ruiz.
\newblock Platypus: Quick, cheap, and powerful refinement of llms.
\newblock \emph{arXiv preprint arXiv:2308.07317}, 2023.

\bibitem[Dettmers et~al.(2022)Dettmers, Lewis, Belkada, and Zettlemoyer]{dettmers2022llm}
Tim Dettmers, Mike Lewis, Younes Belkada, and Luke Zettlemoyer.
\newblock Llm. int8 (): 8-bit matrix multiplication for transformers at scale.
\newblock \emph{arXiv preprint arXiv:2208.07339}, 2022.

\end{thebibliography}

\clearpage

\appendix
\section{Extended Literature Review}
\label{app:extended-lit-review}
In this section, we review practical strategies for post-training efficiency 
and discuss
some
scientific investigations 
that 
provide motivation for, or insight into,
our approach:
in \S\ref{sec:lit-review-pruning}, we first review the history of pruning and then discuss its modern 
application to LLMs;
in \S\ref{sec:lit-review-distil}, we contrast pruning with distillation, an alternative strategy for reducing the parameter count of LLMs;
then in \S\ref{sec:lit-review-PEFT}, we discuss the various practical methods for efficient finetuning and inference acceleration that can be used in conjunction with our pruning strategy;
finally in \S\ref{sec:science-of-dl} we highlight some scientific investigations into
some
depth-dependent statistical properties of LLMs that 
are complementary to our results.

\subsection{Pruning}\label{sec:lit-review-pruning}
\emph{Pruning} is a method for reducing the size of a trained machine-learning model by removing unnecessary parameters, either individually or together as a group.
Pruning for neural networks has a long history \citep{NIPS1989_6c9882bb, NIPS1992_303ed4c6},
and, as originally conceived, 
\emph{unstructured pruning} techniques
sparsify networks by removing individual parameters based on pre-defined criteria. 
For instance, if a parameter of the model has a very small value, then removing it -- i.e. by setting it to exactly zero -- will likely have minimal impact on 
performance.
Inspired by this early work, modern researchers began exploring different criteria for 
such
unstructured
pruning, focusing mostly on computer vision models \citep{han2015learning, chen2015compressing, srinivas2015data}.
In particular, \citet{han2015learning} developed an \emph{iterative pruning} method for alternatively pruning and finetuning a network in order to 
reach better compression ratios and performance.

While these models were smaller, they were not 
necessarily
more efficient:
sparsifying networks by removing individual parameters according to 
a
criterion leads to irregular or pseudorandom sparsification patterns that are difficult to accelerate without specialized hardware or libraries designed for sparsity \citep{li2016pruning}. 
To that end, \emph{structured pruning} techniques were developed to remove irrelevant groups of parameters together, such as particular channels or filters in convolutional networks. 
As this increased their practical relevance, 
researchers then began exploring structured pruning across computer vision \citep{li2016pruning,wen2016learning, hu2016network,he2017channel,huang2018condensenet} and pre-transformer NLP architectures \citep{murray2015auto, see2016compression, kim2016sequence}.

Following 
unprecedented progress in language modeling, recent work
has focused on 
applying structured pruning methods to the Transformer \citep{vaswani2017attention}. 
These studies consider nearly every possible component of the model architecture for elimination, with methods
ranging from dropping attention heads \citep{voita2019analyzing, michel2019sixteen, kim2020fastformers}, to dropping layers \citep{fan2019reducing, zhang2020accelerating, fan2021layer,jha2023large,sajjad2023effect,liu2023comflp}, to pruning hidden states \citep{hou2020dynabert}, 
to rank reducing large weight matrices
\citep{sharma2023truth}, replacing sparse weight matrices with smaller dense ones \citep{ashkboos2024sliceGPT},
to 
many
combinations of the aforementioned groups \citep{xia2022structured, lagunas2021block}.

Of the prior work that also considers transformer layer dropping, most \citep{fan2019reducing, zhang2020accelerating, fan2021layer, xia2022structured, sajjad2023effect}  study BERT-style models \citep{devlin2018bert}, while we consider decoder-only GPT-style models \citep{Radford2019LanguageMA} that are most commonly used for large-scale language modeling and generation. 
BERT-style models are 
naturally
suited for 
understanding tasks 
due to their bidirectional masked language modeling (MLM) objective, 
while GPT-style models are instead 
suited for
generation, 
due to their autoregressive 
objective.
While this divide has been questioned in light of more powerful GPT-style models \citep{zhong2023can}, previous work \citep{ethayarajh2019contextual}
has found significant qualitative differences between BERT and GPT models in terms of the evolution of the layer-wise representation of words. Altogether, this 
suggests that layer-dropping strategies will behave differently between the two families.

One study
for BERT-style pre-trained models, \citet{sajjad2023effect}, concludes that the best layer-pruning strategy is 
dropping the final layers; 
this partially resonates with our results, 
although 
in contrast
we find that 
\emph{(a)} for 
some pruning sizes
keeping the last few layers of the model is actually beneficial, and 
that
\emph{(b)} for all pruning sizes keeping the very last layer is essential.
Additionally, 
while the authors also
study similarity between representations in different layers 
-- as in our approach --
they actually found a higher similarity between representations in the shallow layers compared to the deeper ones -- which 
very sharply disagrees
with our results. 
Importantly, the models considered in \citet{sajjad2023effect} consist of a few hundred million parameters, which is much smaller than the model scales we consider in our work.
Perhaps as a consequence, the authors didn't observe the 
sharp transition
in downstream accuracies that we report in \S\ref{subsec:transition}, despite the fact that they also finetuned their pruned models.

In contrast, while \citet{jha2023large} does consider GPT-style models, the methodology is quite different from ours:  \emph{(i)}  rather than pretraining first and then using a fixed layer-dropping strategy as we do, instead the authors incrementally drop layers in a modified pretraining procedure; and 
\emph{(ii)} the authors study their own sub-1B parameter models, while we focus on the families of readily available, open-weight, large-scale 2.7B-70B parameter models 
that are commonly used and/or finetuned
for practical applications.

As we were finalizing our preprint, \citet{men2024shortgpt} was posted: this paper empirically studies different layer-pruning strategies for GPT-style models (Llama-2 7B and Baichuan2-7B-base) and their subsequent effects on benchmarks (MMLU, CMMLU, and CMNLI). They investigate 
various layer-importance metrics -- notably, their "Block Influence" function is similar to our cosine similarity metric -- and find that they are able to prune up to $\sim$28\% of layers of Llama-2 7B with minimal impact on performance. 
This
provides independent evidence supporting our main takeaway that the deeper layers are not critical for storing knowledge. 

Finally, a systematic approach to layer dropping in transformers has also been 
studied
in the context of \emph{wav2vec} models, 
which are 
encoder-only models that map speech to embeddings and are sized 
in the hundred-million parameter regime \citep{baevski2020wav2vec}.
With these models, \citet{liu2023comflp} developed a layer-pruning algorithm
based on the correlation between layers
and downstream metrics.
Beyond the model architecture and domain, one significant difference between this and our work is that \citet{liu2023comflp} considered non-contiguous pruning proposals, e.g. dropping alternate layers. 
Our intuition for layer pruning predicts that
this shouldn't work as well 
-- at least for decoder-only language models --
as it creates multiple mismatches,
one with each block of layers removed.

\subsection{Model distillation}\label{sec:lit-review-distil}

A completely different
method for reducing the size of a trained machine-learning model is 
\emph{model distillation} \citep{hinton2015distilling}, 
in which knowledge is 
transferred
from a 
large
``teacher'' model 
to a 
smaller 
``student''
model by training the student on the distribution 
predicted
by the teacher.
The essential insight
is 
that this can
transform the very general knowledge and capabilities of the teacher into more streamlined, compressed, and possibly skill-specific representations. 

While a very general technique, in the 
setting of language models,
distillation has been implemented with 
\emph{(a)}
white-box approaches, 
in which the
the student is trained to imitate the teacher's logits \citep{gu2023knowledge} or hidden states \citep{jiao2019tinybert}; as well as with 
\emph{(b)} 
black-box approaches, 
in which the student
only has access to the output tokens generated by the teacher. 
This latter
approach 
broadly 
covers cases where the student is trained on text
that is augmented by the teacher in some way, such as 
by adding
synthetic labels \citep{wang2021want}, generating high quality synthetic text \citep{eldan2023tinystories, li2023textbooks, gunasekar2023textbooks} by providing chain of thought reasoning \citep{fu2023specializing,hsieh2023distilling}, which aims to enhance the student's reasoning skills, or by annotating 
instructions 
that enhance
the student's
instruction-following capabilities \citep{jiang2023lion}. 

Compared to layer pruning, these distillation methods require
considerable computational resources due to the 
reliance on the large teacher to process a big corpus of data.
Instead, our similarity-based pruning strategy only requires computing the similarity between representations at different layers on a small subset of a pretraining corpus, while our second simpler pruning strategy
only 
uses
the reduced model post pruning.

\subsection{Efficient finetuning
and inference acceleration}\label{sec:lit-review-PEFT}

Complementary to directly reducing 
size of a model,
\emph{parameter-efficient finetuning} (PEFT)
focuses on reducing the cost of specializing LLMs to certain tasks.
In particular, Low Rank Adapters (LoRA) reduce the memory and compute of fine tuning by freezing the pretrained model and introducing a parametrically small number of additional trainable weights
\citep{hu2021lora}.
We use its quantized cousin, QLoRA \citep{dettmers2023qlora}, to keep our experiments cost efficient. 
Other PEFT methods that can be combined with our work are ~\citet{li2023loftq} and \citet{zhang2023adaptive}: in the first, the initialization of the LoRA matrices is adjusted to a quantization scheme; in the second, %
LoRA ranks for different LLM modules are chosen in an adaptive manner.

For 
additional
efficiency gains we could 
combine our layer-pruned models with methods that further accelerate inference: with speculative decoding \citep{leviathan2023fast}, tokens are rapidly generated from a smaller draft model and then evaluated in parallel by the main model; with Medusa \citep{cai2024medusa} the draft model is discarded for extra decoding heads, but ultimately achieves a similar effect.
In particular, it could be interesting to consider highly-compressed layer-pruned models as 
potential
draft models in a speculative decoding setup.

\subsection{A breadth of depth-dependent studies}
\label{sec:science-of-dl}

Finally, let us
highlight some scientific work that 
study the depth-dependent properties of LLMs.
One relevant direction 
considers
how
knowledge and linguistic properties are encoded in 
language models.
On the one hand,  \citet{meng2022locating} and \citet{dai2021knowledge} analyze the 
\emph{storage
and recall} 
of factual associations: these works
emphasize that knowledge localizes within the middle \citep{meng2022locating} or final \citep{dai2021knowledge} layers, 
which has implications for
directly
editing or erasing part of a model's factual knowledge.
On the other hand,
attempts to perform such editing gives
evidence that information 
may be
stored non-locally across layers \citep{hase2023does}.
Relatedly,
\citet{geva2023dissecting} investigates
the way
facts are \emph{processed} 
during inference,
distinguishing between the role of attention heads, 
for attribute extraction,
and the MLP blocks, 
for subject enrichment:
both 
are delocalized across several layers.

Next, following the earlier ``logic lens'' \citep{logit_lens_2020},
\citet{belrose2023eliciting} 
invented a technique they called ``tuned lens'' to
study the \emph{trajectory of predictions} by
using a learnable affine transformation to convert intermediate representations into a distributions over tokens (see also \citet{din2023jump}).
By studying the layer-to-layer dynamics of this distribution, the authors noted that it tended to converge. 
This convergence is very suggestive that
that the deeper layers could be prunable, while the fact that they had to train an affine probe is likely related to our observation that the final layer cannot be pruned.
Somewhat relatedly,
\citet{gurnee2023language}
observed that 
geographic features in the underlying text
can be determined from linear probes trained on
intermediate activations,
as long as the activations are deeper than halfway.

More abstractly,
\citet{voita2023neurons} and \citet{liu2023deja} 
found that
the sparsity of activations 
transitions
at around halfway through a network's forward pass,
evolving from sparse to dense.
Perhaps relatedly, 
\citet{panigrahi2023task}
investigated
which model weights update the most during finetuning,
finding that 
it's those in the mid-layers.

Altogether,
these deep studies are complementary to our work, which, 
on the one hand,
provides evidence that removing the deepest layers of an LLM
does not significantly alter the model's performance, 
and,
on the other hand,
demonstrates
a sharp pruning transition after removing approximately half of an LLM's deepest layers.

\section{Experimental Details}
\label{app:healing-procedure}
Here
we explain various 
details
of 
models 
and healing
(\S\ref{app:finetuning-details}) %
and of evaluations (\S\ref{app:evaluation_details}).

\subsection{Model and healing details}
\label{app:finetuning-details}
All models in this paper were fine-tuned using the Hugging Face \texttt{Trainer API} \citep{wolf_etal_2020_transformers}.
A list of models and their paths on Hugging Face
are as follows:
\begin{center}
\begin{tabular}{l ||l}
\toprule
\textbf{Model} & Repository Path \\
\midrule
Llama-2 7B & \verb|meta-llama/Llama-2-7b-hf|
\\
Llama-2 13B & \verb|meta-llama/Llama-2-13b-hf|
\\
Llama-2 70B & \verb|meta-llama/Llama-2-70b-hf|
\\
Mistral 7B & \verb|mistralai/Mistral-7B-v0.1| \\
Phi-2 (2.7B) & \verb|microsoft/phi-2| \\
Qwen 7B & \verb|Qwen/Qwen-7B| \\
Qwen 14B & \verb|Qwen/Qwen-14B| \\
\end{tabular}
\end{center}

For healing,
we
used
the version of the Colossal Clean Crawled Corpus (C4) \citep{2019t5} from Hugging Face:  \verb|data = load_dataset("c4", 'en')|.
We truncated long examples as described 
later in the
paragraph and added special tokens when available.\footnote{N.B. the Qwen tokenizer from Hugging Face does not include any special tokens; in this case, it was essential to add a default padding token.
}  
Models were finetuned
for 5000 steps with
a global batch size of 16: this corresponds to total finetuning tokens of $16 \times 5000 \times [\text{\texttt{max\_seq\_length}}]$ for each model.
We used
a cosine-annealed learning rate schedule, with a warmup of 100 steps.
When possible, the peak learning rate was 
set to the peak learning rate from the model's pretraining;
in practice, 
this 
means all models were 
trained with a peak LR of 3e-4, with the exceptions of Phi-2 \citep{phi2}, which was trained with a peak LR of 2e-4 during pre-training, Llama-2-70B, which 
was trained with a peak LR of 3e-5 (a value that resulted from a sweep), and Mistral-7B which was trained with a peak LR of 3e-6 (also a value that resulted from a sweep).
All models 7B parameters or smaller
were trained with a max sequence length of 2048 tokens, while
all models
13B parameters or greater
were trained with a max sequence length of 4096 tokens.
While we realize that some models may have been pretrained on longer
sequences, e.g. Qwen\emph{-the-outlier} \citep{bai2023qwen}, 
we decided to 
the max sequence length consistent across models of similar size to allow fairer comparisons across model families.

On top of the 
Hugging Face Trainer API, 
we used quantization and Low-Rank Adapters (LoRA) \citep{hu2021lora} for all of our finetuning:
\begin{itemize}
\item For quantization, we used the \texttt{bitsandbytes} library for QLoRA~\citep{dettmers2023qlora}
to
quantize our models to 4 bits. 
\item For LoRA, we used the Hugging Face \texttt{peft} library \citep{peft}. We set the LoRA dropout to 0.05 and kept the LoRA $\alpha$ equivalent to the LoRA rank, following \citep{lee2023platypus}. Aside from two exceptions, 
discussed below, models are trained with LoRA rank 64.
\item Also following \citet{lee2023platypus}, we only applied LoRA to FFN modules:  \verb|["gate_proj", "down_proj", "up_proj"]| for Llama-2 and Mistral models, \verb|["fc1", "fc2"]| for Phi-2, and \verb|["w1", "w2", "c_proj"]| for Qwen models.%
\end{itemize}

The large majority of these hyperparameter choices are standard and found in previous works, e.g. \citet{lee2023platypus} and \citet{dettmers2022llm}. For absolute clarity, we list display all the model specific architecture and healing details below:
\begin{center}
  \begin{tabular}{l| c r c c c r }
  \toprule
    \textbf{Model} & \# Layers & Vocab Size & Max Seq. Len. & FT Tokens & Peak LR & LoRA Rank \\
    \midrule
    Llama-2 7B & 32 & 32,000 & 2048 &164M &3e-4  & 2 \\
    Llama-2 13B & 40 & 32,000 & 4096 &328M &3e-4 &  64 \\
    Llama-2 70B & 80 & 32,000 & 4096 &328M &3e-5 &  8 \\
    Qwen 7B & 32  & 151,936 & 2048 &164M &3e-4 &  64 \\
    Qwen 14B & 40  & 151,936 & 4096 &328M &3e-4 &  64 \\
    Mistral 7B & 32  & 32,000 & 2048&164M & 3e-6 &  4 \\
    Phi-2 2.7B & 32 & 51,200 & 2048&164M & 2e-4 &  64 \\
\end{tabular}

  \end{center}
We also have the following hyperparameters common between all models:
\begin{center}
  \begin{tabular}{l |l }
  \toprule
    \textbf{Config} & Value \\
    \midrule
    Finetuning dataset & C4\\
    Batch size & 16 \\
    LoRA $\alpha$ & LoRA rank \\
    LoRA dropout & 0.05 \\
    LoRA targets & FFN modules \\
    LR scheduler & Cosine \\
    Warmup steps & 100 \\
    Total steps & 5000 \\
  \end{tabular}
\end{center}

\subsection{Evaluation details}
\label{app:evaluation_details}
We performed three 
principal
evaluations:
accuracy on \emph{MMLU}, accuracy on \emph{BoolQ}, and 
loss on \emph{C4}.

For \textbf{MMLU accuracy}:
\begin{itemize}
\item We use the \verb|cais/mmlu| version of the dataset from Hugging Face. 
\item We 
follow 
the formatting suggested in the original reference \citep{hendrycks2020measuring} 
without further prompt engineering.
\item For constructing few-shot examples, we use the \verb|dev| set from \verb|cais/mmlu|.
\item For our experiments, we use $0$ few-shot examples; our results and analysis are robust to this choice, cf. Figure~\ref{fig:appendix_effect_prompt_change}.
\item We report average accuracy across all subjects.
\end{itemize}

\begin{figure}[th]
\begin{center}
\centerline{\includegraphics[width=1.0\columnwidth]{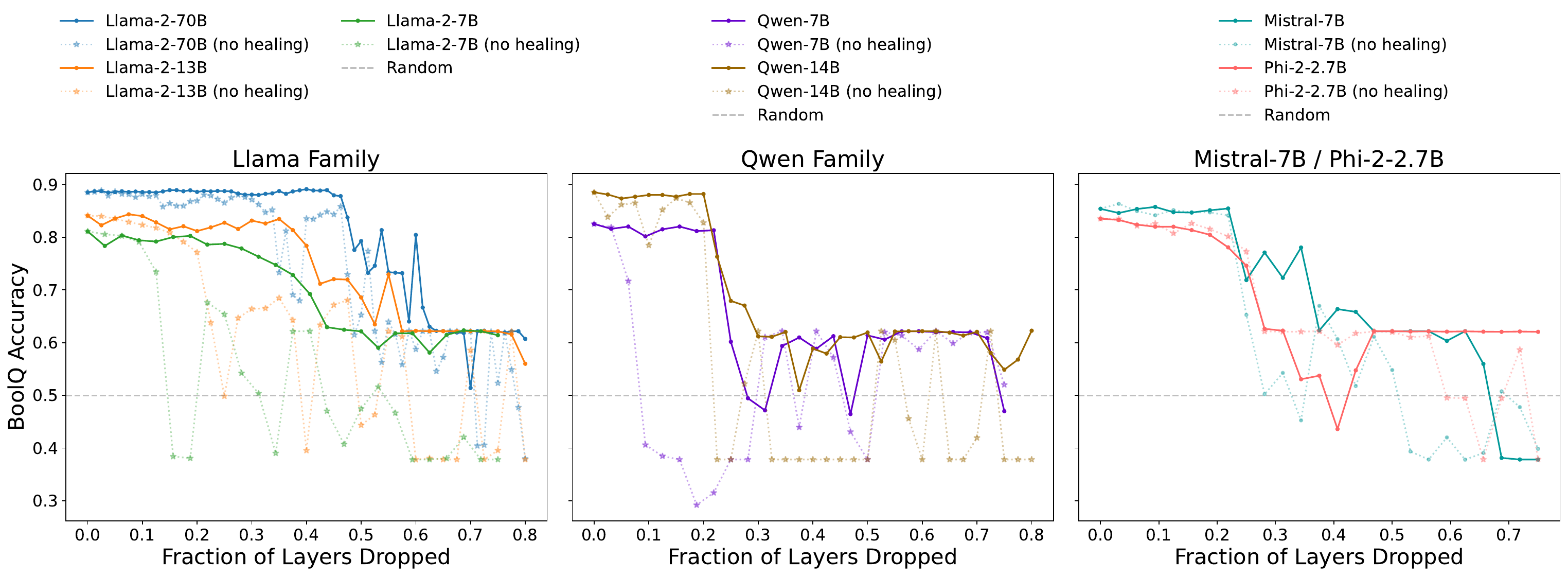}}
\caption{BoolQ accuracy (0-shot) 
vs. fraction of layers 
dropped
for different model families. 
(\emph{Left:} Llama-2 family; \emph{Middle:} Qwen family; \emph{Right:} Mistral-7B and Phi-2.)
The 
solid lines
represent 
performance
after dropping layers and healing, 
and the (semi-transparent) dotted lines
show 
performance after dropping layers only (no healing),
and the dashed gray line is the score for guessing randomly.
For BoolQ, healing leads to important improvements such that performances; then, across all models,  performances
are quite robust 
until 20\%-55\% pruning fractions, depending on model family and size, at which point they transitions to random guessing.
}

\label{fig:appendix_boolq_acc}
\end{center}
\end{figure}

For \textbf{BoolQ accuracy}: 
\begin{itemize}
\item We used the \verb|hassansh/boolq_n_shot| version from Hugging Face. 
\item For our experiments, we use $0$ few-shot examples.
\item The complete BoolQ results -- truncated from the main text -- are shown here in Figure~\ref{fig:appendix_boolq_acc}: in the left panel we present the Llama-2 family, in the middle panel we present models from the Qwen family, and in the right panel we should Mistral-7B and Phi-2; we also make the experiments without healing semi-transparent in order to better display the results from the complete similarity-informed pruning method. Importantly, while we see here that healing plays a more important role than it did 
for MMLU in Figure~\ref{fig:main-results-pruning}, after healing we still have a characteristic flat region of robust performance; 
as before, the capabilities required to achieve a model's top score isn't removed by significant layer pruning until a critical model-dependent threshold.
\end{itemize}

For \textbf{C4 Validation Loss}: 
\begin{itemize}
\item We used the \verb|c4| version from Hugging Face (soon be deprecated in favor of \verb|allenai/c4|).
\item We evaluated using the \emph{validation} split as we healed with the train split.
\item Given its size, we randomly sampled 60k sequences and held them fixed across all models.
\item In Figure~\ref{fig:c4-validation-plots} we normalized the 
loss to facilitate fair comparison across model families that employ different vocab sizes:
to normalize, we divided by
$\log V$, where $V$ is the \emph{per-model} vocab size (listed in a table in \S\ref{app:finetuning-details}). This, $\log V$, corresponds to the loss of sampling tokens uniformly, which naturally sets the scale for a given model.
\end{itemize}

\section{Ablations}
\label{app:hyperpara}
Here
we detail various ablations: 
prompting (\S\ref{app:prompt}), finetuning seed (\S\ref{app:fine-tuning-seed}),
LoRA rank (\S\ref{app:lora-rank}),
other pruning strategies (\S\ref{subsec:other-pruning-methods}).
Qualitatively, the results of the paper are quite robust to the variation of any of these.

\subsection{Prompting}
\label{app:prompt}
\begin{figure}[th]
\begin{center}
\centerline{\includegraphics[width=1.0\columnwidth]{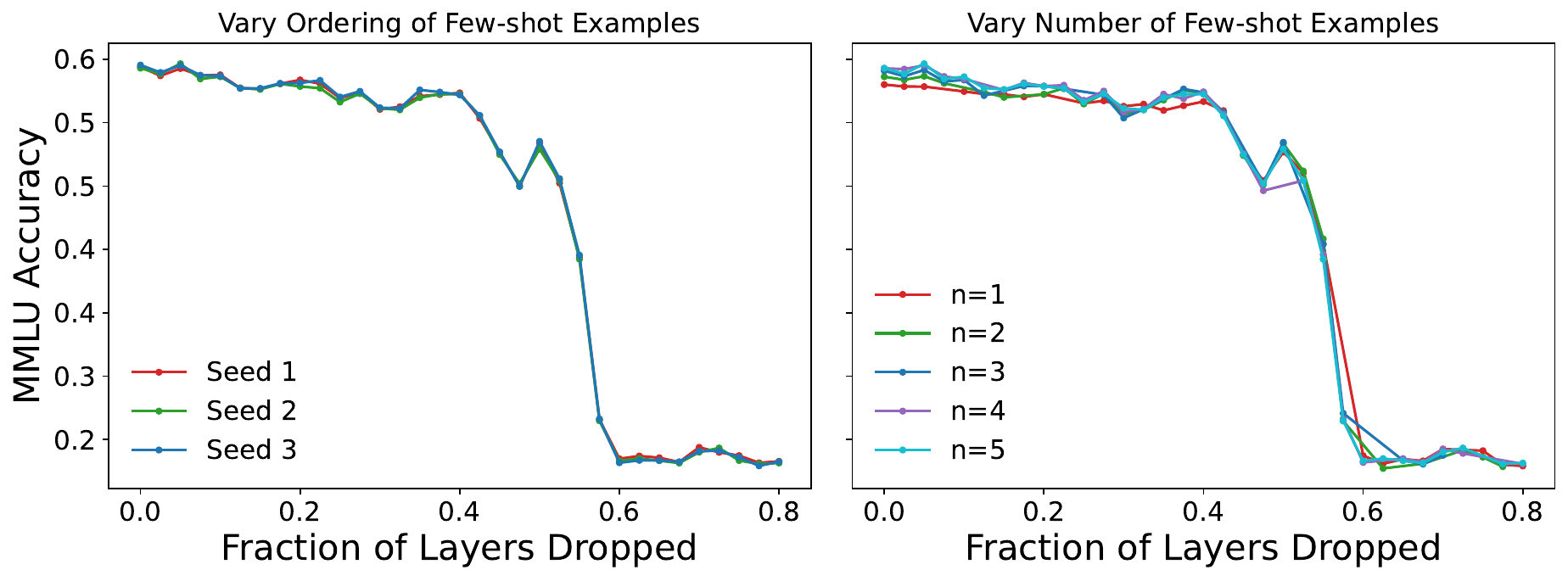}}
\caption{Effect of prompt ablations on MMLU accuracy vs. fraction of layers dropped for
Llama-2-13B.
\emph{Left:} We vary the ordering of the few-shot examples and see it does not have any impact.
\emph{Right:} We very the number $n$ of few-shot examples; while careful study of the flat region suggests increasing the number of few-shot examples marginally improves performance,
regardless,
the layer-pruning strategy is robust to 
this kind of variation.
}

\label{fig:appendix_effect_prompt_change}
\end{center}
\end{figure}

It's common knowledge that altering the prompt on QA evaluations can significantly impact results. To control for prompting,  we ablate the MMLU accuracy 
for our principal similarity-informed pruning described in \S\ref{subsec:layer-pruning-algo} when applied to
Llama-2-13B:
in the left panel of Figure~\ref{fig:appendix_effect_prompt_change}, we show results for
changing the ordering of the few-shot examples in the prompt, 
and in the right panel the same figure, we show results for changing the number of few-shot examples. Broadly we see that the layer-pruning method is robust to these changes.

\label{app:ablations}
\subsection{Finetuning seed}
\label{app:fine-tuning-seed}
\begin{figure}[th]
\begin{center}
\centerline{\includegraphics[width=0.6\columnwidth]{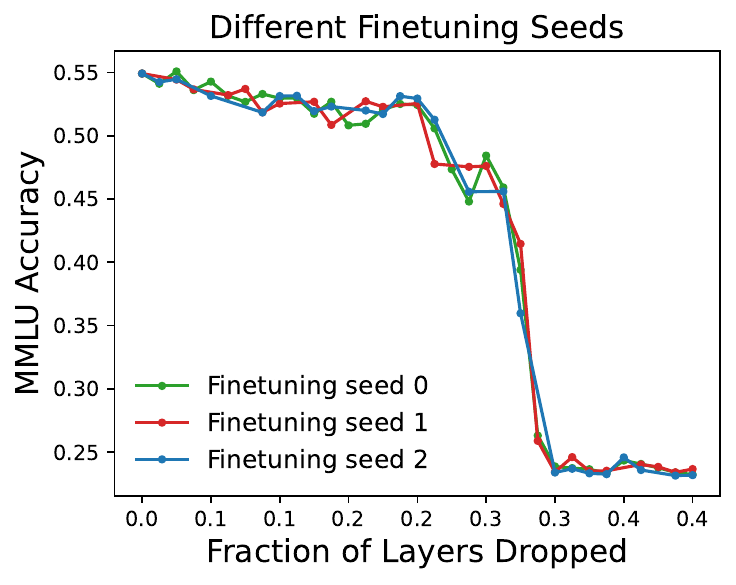}}
\caption{Effect of varying the finetuning seed on MMLU accuracy vs. fraction of layers dropped for Llama-2-13B: there is no meaningful effect.
}

\label{fig:appendix_effect_of_finetuning_seed}
\end{center}
\end{figure}

Here we vary the finetuning seed.
For all of our experiments, we use the following code snippet to ensure reproducibility:
\begin{verbatim}
SEED_VAL = 0
transformers.enable_full_determinism(SEED_VAL)
\end{verbatim}
Since we begin with a pretrained model, the finetuning seed doesn't affect initialization, but it will impact the stochastic aspects of further training such as data order. 
To control for this,  we ablate the 
finetuning seed
for our principal similarity-informed pruning described in \S\ref{subsec:layer-pruning-algo} when applied to
Llama-2-13B: in Figure~\ref{fig:appendix_effect_of_finetuning_seed} we observe that the layer-pruning method is robust to the choice of seed.

\subsection{LoRA rank}
\label{app:lora-rank}
Here we vary the LoRA rank used for healing. Unfortunately, our compute budget did not allow us to make an exhaustive sweep across all of our experimental configurations. In lieu of that, we employed the following protocol for our main experiments:
\begin{itemize}
\item Begin with rank 64,
following the QLoRA setup (see, e.g. Appendix B.2 of \citet{dettmers2023qlora}).
\item If healing with that rank significantly harms the performance compared to no healing, then 
sweep LoRA ranks for that model and, for the other evaluations, pick the best performing LoRA rank according to its MMLU accuracy.
\end{itemize}
This protocol is designed to maximize the chance that healing will improve performance across all of our evaluations. For simplicity, we ran this rank-picking protocol using the simple pruning heuristic, with the exception of Llama-2-70B.

\begin{figure}[th]
\begin{center}
\centerline{\includegraphics[width=1.0\columnwidth]{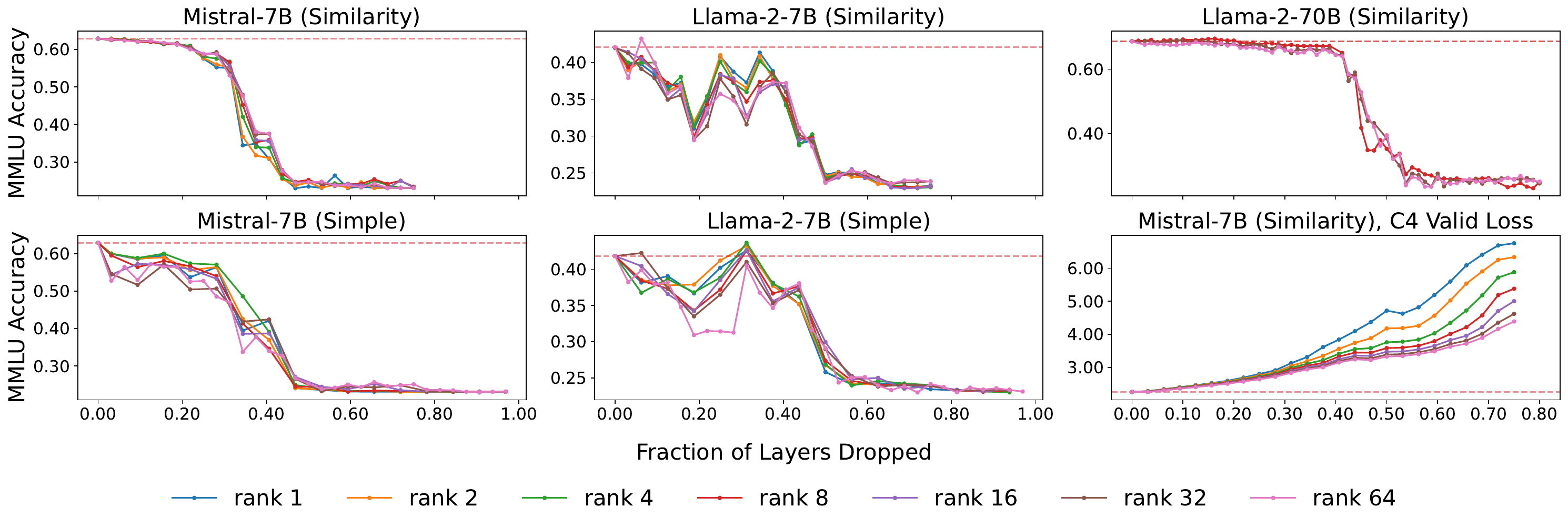}}
\caption{Effect of varying the LoRA rank. 
\textbf{Top}: 5-shot MMLU accuracy vs. fraction of layers dropped 
using the similarity-informed pruning strategy on
Mistral-7B (\emph{left}), Llama-2-7B (\textit{middle}), and Llama-2-70B (\textit{right}).
Across all ranks we observe 
similar behavior, though 
there's a small effect of
decreasing
rank 
improving
overall performance. 
\textbf{Bottom, left and middle}: 5-shot MMLU accuracy vs. fraction of layers dropped 
using the simple pruning heuristic on
Mistral-7B (\emph{left}) and Llama-2-7B (\textit{middle}). As before, qualitative behavior is similar across ranks, though 
in this case 
it's much clearer that decreasing rank improves performance.
\textbf{Bottom, right}: C4 validation loss vs. fraction of layers dropped
using the similarity-informed pruning strategy on
Mistral-7B.
In contrast to MMLU, decreasing rank 
harms performance; together, these results suggest that larger ranks may be overfitting.
}
\label{fig:appendix_effect_of_rank_lora_sweep}
\end{center}
\end{figure}

In practice, this led to us using rank 64 for every model with the exceptions of 
Mistral-7B, with rank 4,
Llama-2-7B, with rank 2,
and
Llama-2-70B, with rank 8. 
(To review 
this same information 
in tabular form,
see the second Table in \S\ref{app:finetuning-details}.)
Figure~\ref{fig:appendix_effect_of_rank_lora_sweep}
displays the sweeps over MMLU accuracy
supporting these choices 
for Mistral-7B (bottom left panel), 
Llama-2-7B (bottom middle panel),
and 
Llama-2-70B (top right panel):
overall, while the LoRA rank does not have a significant impact on the qualitative behavior of the healed model,
decreasing the LoRA rank 
generally 
improves performance. In the top left and middle panels of Figure~\ref{fig:appendix_effect_of_rank_lora_sweep}, we show corresponding sweeps for Mistral-7B (top) and Llama-2-7B (middle) using the similarity-informed pruning strategy:
we see that for this pruning method both models are much more robust,
though rank 2 is still the top performing rank for Llama-2-7B.

The characteristic improvement of MMLU accuracy with decreasing LoRA rank -- even for extremely low ranks(!) -- deserves an explanation. One possibility is that lowering the LoRA rank can better regularize finetuning against overfitting. In particular, astute readers may have been surprised at the discussion of peak learning rates in \S\ref{app:finetuning-details}: models were finetuned with the same peak used in pretraining; 
a ``large'' LoRA rank of 64 introduces a number of additional parameters that may overfit to C4. This overfitting would certainly be harmful, since 
the actual pretraining datasets for the models we consider are \emph{(a)} unknown to us, and \emph{(b)}, likely to be of significantly higher quality than C4.

We investigate this directly for Mistral-7B. In the bottom right panel of Figure~\ref{fig:appendix_effect_of_rank_lora_sweep} we plot the C4 validation loss
across different LoRA ranks: we see that while decreasing the LoRA rank generally improves MMLU accuracy (cf. left-most panels), at the same time it harms the C4 validation loss. 
This supports our overfitting hypothesis. In a greater-resourced future, it would be interesting to improve the healing process by considering other forms of regularization and learning rate tuning.

\subsection{Other pruning strategies}
\label{subsec:other-pruning-methods}
\begin{figure}[th]
\begin{center}
\centerline{\includegraphics[width=0.6\columnwidth]{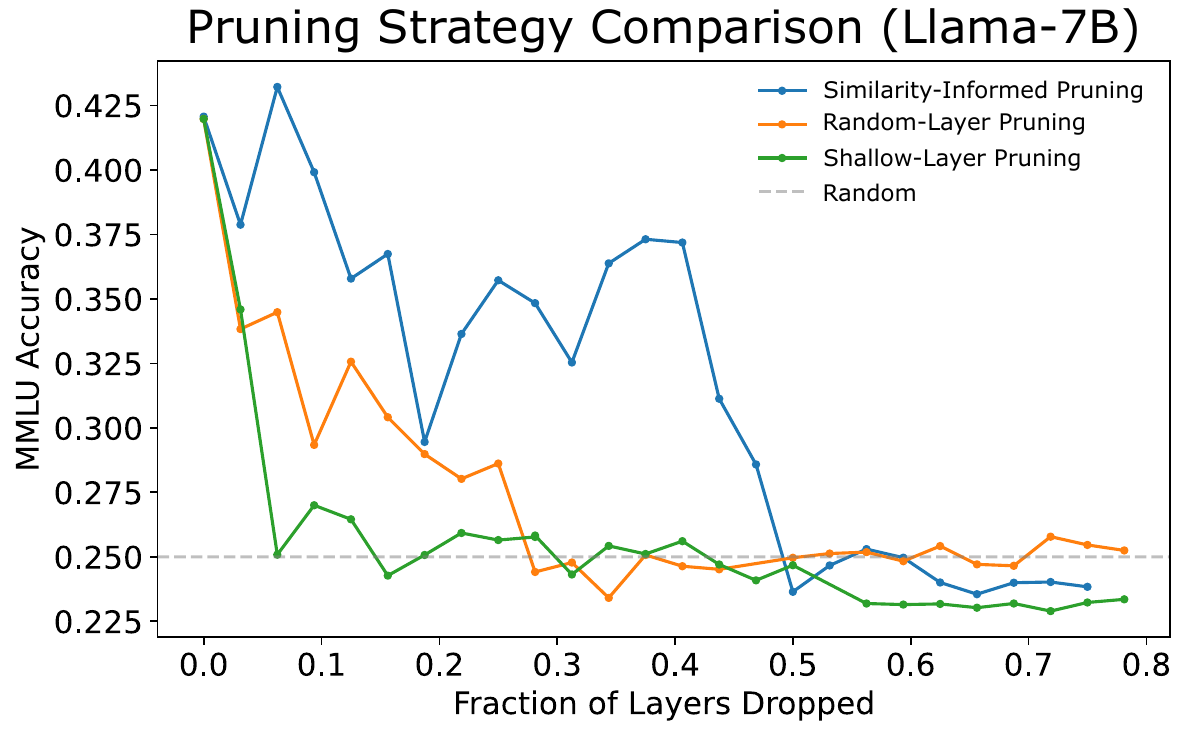}}
\caption{Comparison of the similarity-informed pruning strategy (blue) to random-layer pruning (orange) and shallow-layer pruning (green) on MMLU accuracy, with Llama-2 7B and LoRA rank 64. The similarity-informed pruning strategy clearly outperforms these baselines.
}

\label{fig:appendix_comparison_to_other_pruning_methods}
\end{center}
\end{figure}

Here we study how the similarity-informed pruning strategy (\S~\ref{subsec:layer-pruning-algo}) compares to other layer-pruning baselines: specifically, we contrast with pruning random layers and pruning shallow layers. In Figure~\ref{fig:appendix_comparison_to_other_pruning_methods}, we observe that the similarity-informed strategy from the main text outperforms both of these other strategies on an MMLU evaluation of Llama-7B.

\end{document}